# Shaping AI's Impact on Billions of Lives

Mariano-Florentino Cuéllar, Jeff Dean, Finale Doshi-Velez, John Hennessy, Andy Konwinski, Sanmi Koyejo, Pelonomi Moiloa, Emma Pierson, and David Patterson

# Introduction

Artificial Intelligence (AI), like any transformative technology, has the potential to be a double-edged sword, leading either toward significant advancements or detrimental outcomes for society as a whole. As is often the case when it comes to widely-used technologies in market economies (e.g., cars and semiconductor chips), commercial interest tends to be the predominant guiding factor. The AI community is at risk of becoming polarized to either take a laissez-faire attitude toward AI development, or to call for government overregulation. Between these two poles we argue for the community of AI practitioners to consciously and proactively work for the common good. This paper offers a blueprint for a new type of innovation infrastructure including 18 concrete milestones to guide AI research in that direction. Our view is that we are still in the early days of practical AI, and focused efforts by practitioners, policymakers, and other stakeholders can still maximize the upsides of AI and minimize its downsides.

To offer a sufficiently-broad and realistic perspective that captures the possibilities, we've assembled a team composed of senior computer scientists, policymakers, and rising stars in AI from academia, startups, and big tech—a team that covers many AI domains (see Authors).

In addition to our own expertise, our perspective is informed by interviews with two dozen experts in various fields. We talked to luminaries such as recent Nobelist John Jumper on science, President Barack Obama on governance, former UN Ambassador and former National Security Advisor Susan Rice on security, philanthropist Eric Schmidt on several topics, and science fiction novelist Neal Stephenson on entertainment. We also met with experts in labor economics, education, healthcare, and information. This ongoing dialogue and collaborative effort has produced a comprehensive, realistic view of what the actual impact of AI could be, from a diverse assembly of thinkers with deep understanding of this technology and these domains.

> *Our view is that we are still in the early days of practical AI, and that focused efforts by practitioners, policymakers, and other stakeholders can still maximize the upsides of AI and minimize its downsides.*

These discussions have crystallized our conviction that recent AI models have shown a remarkable promise to influence the world, potentially affecting billions of lives for better or worse. We think the best bet going forward is to assume AI progress will continue or speed up, and not slow down. AI's impact on society will be profound.

From these exchanges, five recurring guidelines emerged, which form the cornerstone of a framework for beginning to harness AI in service of the public good. They not only guide our efforts in discovery but also shape our approach to deploying this transformative technology responsibly and ethically.

**1. Humans and AI systems working as a team can do more than either on their own.** Applications of AI focused on *human productivity* produce larger productivity increases than those focused on *replacing human labor* [Brynjolfsson] [National Academies]. In addition to increasing people's employability, tools aimed at making people more productive let them act as safeguards if AI systems veer off course. AI at times can level the playing field between those who have many resources and those of limited resources. Since people and AI systems tend to make different mistakes, collaborating with AI may improve results. In short, focusing on human productivity helps both people and AI tools succeed. Policies should aim toward innovations that encourage human-AI collaboration while reducing risks.

**2. To increase employment, aim for productivity improvements in fields that would create more jobs.** Despite tremendous productivity gains in computing and airline travel, the United States in 2020 had 11 times more programmers and 8 times more commercial airline pilots than in 1970. This growth is because programming and airline transportation were fields with what labor economists call an *elastic* demand. Goods with elastic demand are those where a decrease in price results in a large increase in the quantity acquired. Agriculture, on the other hand, is *inelastic* in the U.S., so productivity gains have reduced the number of agriculture jobs fourfold in one human lifetime (1940 to 2020). Discussions with experts in other fields will likely uncover more opportunities for AI to increase productivity. If policymakers and practitioners aim AI systems at improving productivity in elastic fields, AI can *increase* employment, despite public fears to the contrary. And as recent Nobelist John Jumper observed, one way to accelerate scientific progress is to improve the productivity of *scientists*, which is the goal of a "scientist's aide" (see Science). Productivity gains in science from AI could prove to be extremely valuable to society [National Academies].

**3. AI systems should initially aim at removing the drudgery of current tasks.** If policymakers and practitioners first target AI systems that automate menial and unfulfilling aspects of current jobs, they can make work more meaningful and enjoyable. Doctors and nurses choose their careers because they want to help patients, not to do endless insurance documentation. Schoolteachers may prefer to spend their time on student interaction rather than grading and recordkeeping. Rather than skip ahead to new AI innovations, first provide AI tools to improve the meaningfulness of people's current work in hospitals and classrooms. For example, AI-powered "teacher's aide" tools (see Education) could automate tasks teachers find unfulfilling, freeing up time to spend with students and making teaching workloads more manageable. A secondary benefit is that they might be more likely to use AI tools in the future.

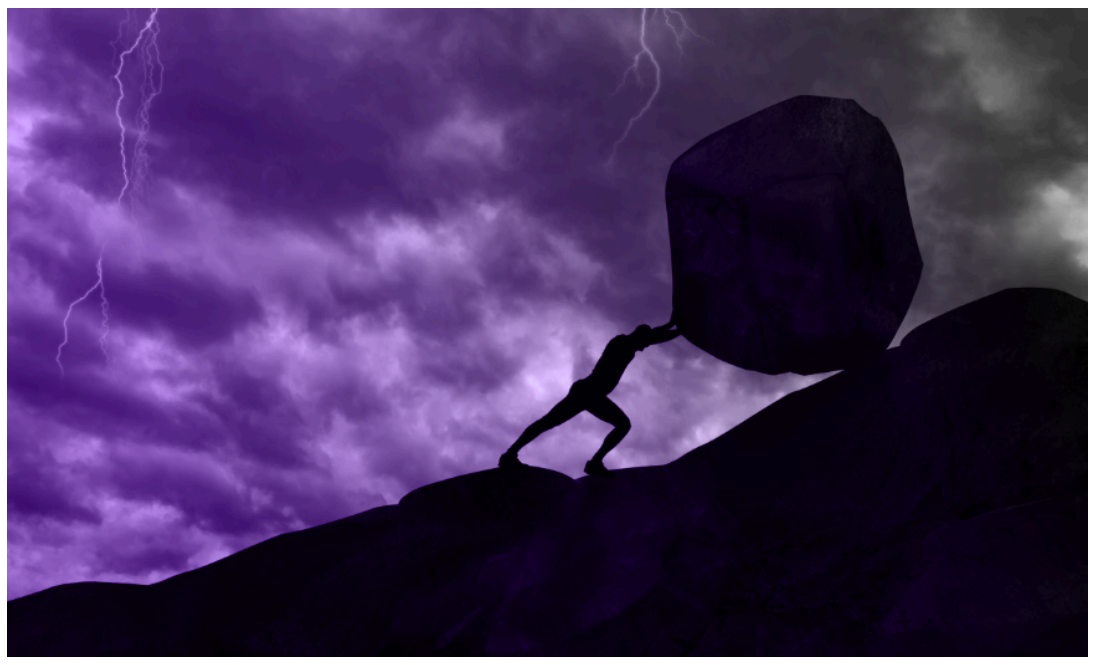

**Sisyphus's drudgery. He could have used AI's help.**

**4. The impact of AI varies by geography.** Philanthropist Eric Schmidt emphasized that while nations with advanced economies worry about AI displacing highly trained professionals, countries with lean economies face shortages of these same skilled experts (see Employment). AI could make such expertise more widely available in places with extreme scarcity of trained workers and with insufficient funding, potentially enhancing quality of life and economic growth. AI systems could become as transformative for the low- and middle-income nations as mobile phones have been [Rotondi]. For example, a "healthcare aide" that improved the skill sets and productivity of nurses and physician assistants could also give more patients access to quality healthcare in regions facing physician shortages (see Healthcare). Multilingual AI models on smartphones can greatly help people in low- and middle-income countries gain access to information, education, media/entertainment, and more. Better economies and services may even offer alternatives to emigration for some in middle-income countries.

**5. Determine the best metrics and methods to evaluate AI innovations.** We must measure AI accurately to evaluate its real potential. In high-stakes domains, because we can't risk harming participants, we need to use gold standard tools to evaluate innovation accurately and identify possible limitations before wide deployment: *A/B testing*, *randomized controlled trials*, and *natural experiments*.[1] Equally urgent is post-deployment monitoring to evaluate whether AI innovations do what they say they are doing, whether they are safe, and whether they have externalities. We also need to continuously measure AI systems in the field so as to be able to incrementally improve them. In other, lower risk situations, the marketplace and observational studies can assess effectiveness of AI tools without needing the same rigor, such as for AI tools for programmers.

Having covered the five guidelines, the next part sets the context for the current excitement about AI.

# I. Putting Pragmatic AI in Context

## History of Technological Paradigm Shifts

Similar to the dawn of television, computers, nuclear power, and the internet, uncompromising antagonistic positions are being taken in these early days of practical AI. The polarized discourse on this new technology has devolved currently into a standoff between "accelerationists" and "doomers." Like most practitioners of AI, we believe reality is more nuanced.

One debated issue is the role of the government in AI's development. Recent efforts by companies to develop AI systems have been likened to the Manhattan Project in the 1940s or the Space Race of the 1960s. In terms of investment size, the nearly $2B

[1] A *natural experiment* is a research study where individuals are exposed to different conditions, like a control group, not by the researcher's design but by a naturally occurring event or policy change. Researchers treat such a study as acting as if random assignment occurred, allowing them to observe and analyze the effects without actively manipulating variables. This option is often used when controlled experiments are not feasible due to ethical or practical limitations.

for the Manhattan Project would be $27B in today's dollars, and the $26B to put a person on the moon would be $318B today. While current AI is roughly comparable in terms of size of investment, the big difference is that the U.S. government funded those efforts while private industry backs this one, and most of the talent involved are in the AI industry.

Given this relationship, we need a new innovation infrastructure. Policy changes to improve the impact of AI are likely best accomplished via collaboration between government, industry, and academia.[2] As a historical precedent, we can look at the role the government played in the development of integrated circuit chips and cars.

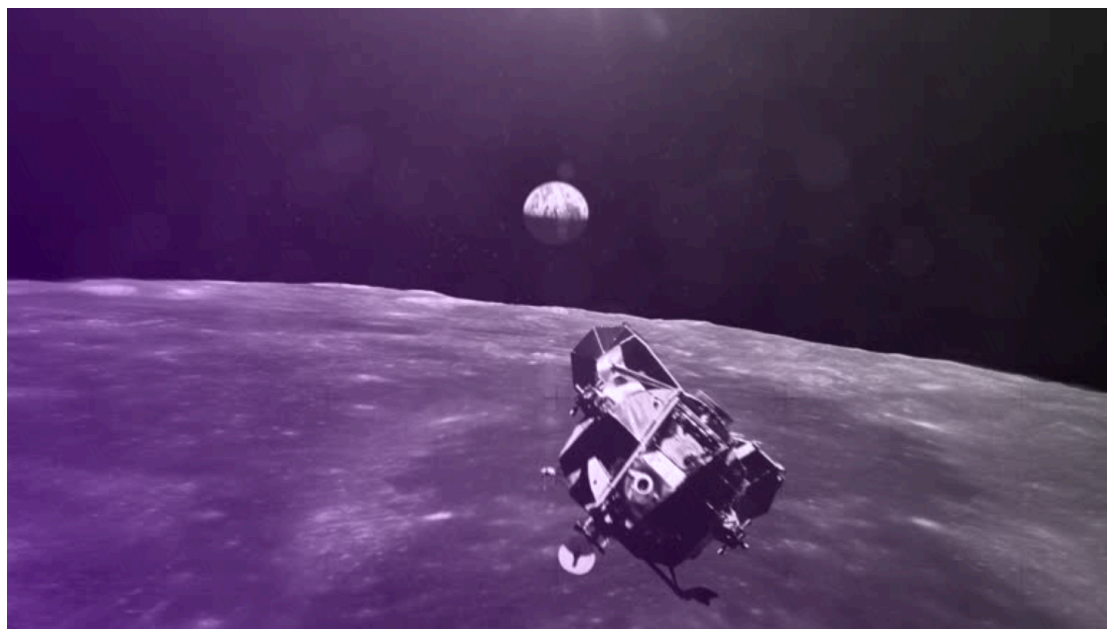

**The U.S. government's Apollo and Minuteman programs used >95% of all chips made in 1965.**

In the 1960s, the government was the primary consumer of chips, as the smaller size and lower power of chips was vital in the Space Race. Over 95% of the chips made in 1965 were used by the Apollo and Minuteman programs. This manufacturing volume allowed the nascent semiconductor industry to improve its fabrication prowess so it could enter the much larger commercial market by the end of the decade. Two years later, Intel delivered the first microprocessor. The government also funded university research that helped push the frontiers of chip design and manufacturing, helping Moore's Law to continue for more than 50 years.

In the first half of the 20th century, car manufacturers benefited as governments built and improved roads and freeways funded by gasoline taxes, created traffic lights and travel signs, and licensed drivers. In the 1960s, the U.S. created the National Highway Traffic Safety Administration and the Environmental Protection Agency, which set societal benefiting standards on car safety and emissions for the whole industry that might have been difficult for individual car manufacturers to do on their own. More recently, the government has funded academic research to improve cars. Examples are DARPA's self-driving challenge (won by academic researchers), automotive safety, and fuel efficiency.

We envision a coordinated public-private partnership for AI. Its goal would be to remove bureaucratic roadblocks (e.g., to sharing data), ensure safety, and provide transparency and education to policymakers and the public. In addition to learning from historical precedents for the development of AI systems, we should also learn from the history of how transformative technologies have been deployed.

One lesson learned from the rollouts of paradigm-shifting technologies like broadband internet, cloud, mobile devices, and social media is that their deployment was lengthier than technologists predicted, but their impact was even more widespread. Quoting Bill Gates:

> *One thing I've learned in my work with Microsoft is that innovation takes longer than many people expect, but it also tends to be more revolutionary than they imagine.*

Another lesson is that predictions of technological impact from people in other fields are similarly inaccurate [National Academies]:

> *… commentators and experts of all stripes—social and natural scientists, historians, and journalists—have an almost unblemished record of incorrectly forecasting the long-run consequences of technological innovations.*

A third lesson is that it is often hard to accurately predict the unintended negative side effects until after the technologies were widely deployed, with social networking as the prime example.

[2] In addition to universities helping advance the research frontier, the people in industry and government pursuing AI technology and policy are educated at universities. Enabling universities to prepare individuals to advance AI, as well as to educate the broad population to thrive in a world of ubiquitous AI, is crucial to our shared future.

Time will tell if AI proves to be an exception to these three lessons.

## Artificial Intelligence (AI)

Before we discuss AI's impact within each of our half-dozen fields, let's review how we got here. The term *Artificial Intelligence (AI)* was coined to define *the science and engineering of making intelligent machines* in 1956, only five years after the first commercial computer.[3]

One strand of AI that became popular over the next decades was to create a set of rules of the form "if this happens do that, if that happens do this." The belief was that with sufficiently accurate and large sets of rules, intelligence would emerge. Within the big tent of AI, a contrarian strand did not accept that humans would ever be able to write such a set of rules. They believed that the only hope was to learn the rules from the data. That is, *it was much harder to program a computer to be clever than it was to program a computer to learn to be clever*. Just three years after AI was defined, they christened this bottom-up approach *machine learning (ML)*.[4]

One branch of the ML community believed the only hope for creating a program that could learn from data would be to imitate our one clear example of intelligence: the human brain. Our brains consist of 100 billion neurons with 100 trillion connections between them. This version of ML is based on a very simple model of a neuron, so this form of ML (that is also within the big tent of AI) is called a *neural network*. A typical neural network might use 100 million artificial neurons. Because current versions of neural networks have many more layers of artificial neurons than in the past, recent incarnations are also called *deep neural networks* or *deep learning*.

Neural networks have two phases, *training* and *serving* (also called *inference*). Training is analogous to being educated in college and serving is like working after graduation. Training a neural network involves repeatedly showing it labeled data (e.g., images identified as cats or dogs) with the system adjusting its artificial neurons until it gives sufficiently accurate answers to questions about that data (e.g., is it a cat or a dog). Once trained, the goal is that the model should work well with data it has not yet seen (e.g., correctly determining if an image is of a cat or a dog).

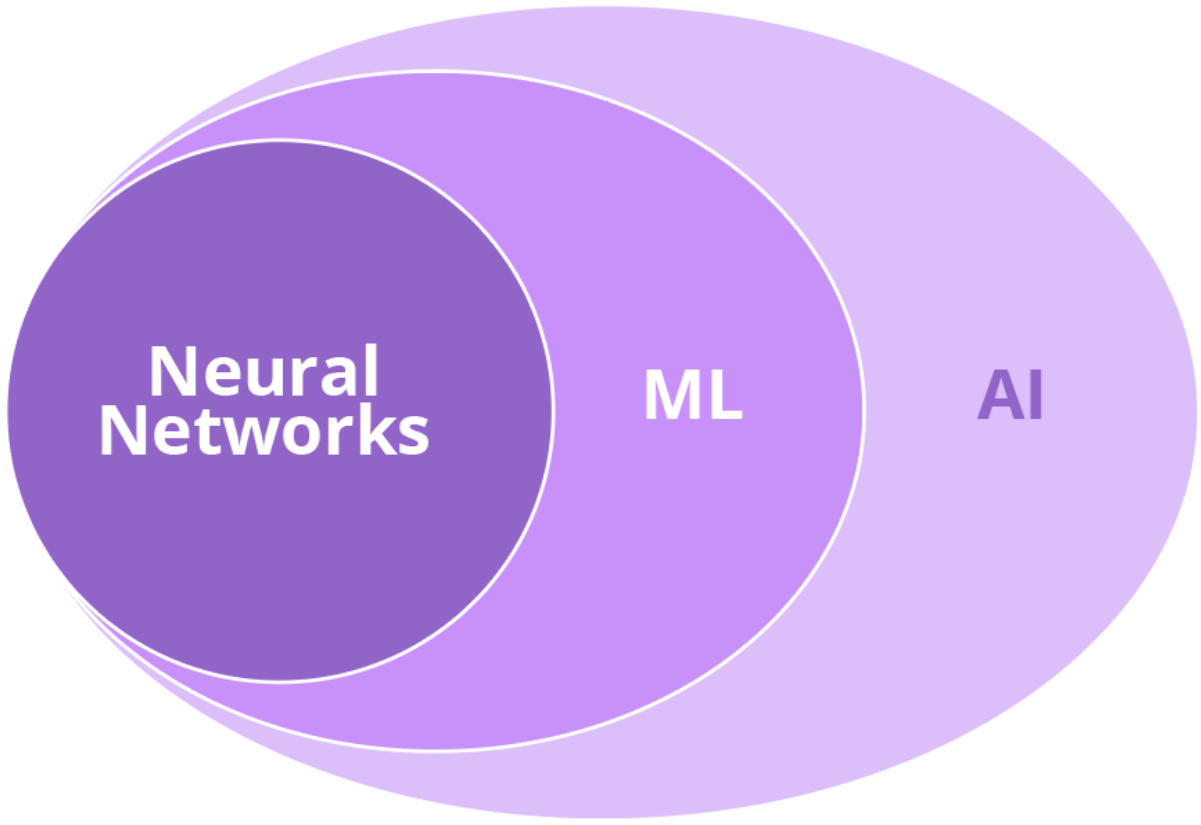

**The Russian dolls of AI.**

After decades of debates about which AI philosophy was best, in 2012 neural networks started to soundly beat the competition. The breakthrough 12 years ago wasn't so much the invention of new neural network algorithms as it was that Moore's Law led to machines that were 10,000 times faster and we could network many together to work in concert. That enabled training using 10,000 times more labeled data available from the World Wide Web. Virtually all news stories today concerning AI breakthroughs are, more precisely, about neural networks.

The excitement about AI spiked by ChatGPT in 2022 is about models with billions of neurons that take months to train on tens of thousands of chips designed solely for neural network training. These giant neural networks were initially called *large language models* (*LLMs*) because the first examples performed amazing feats based on text. Eventually these models became more multimodal, incorporating data types beyond text such as images, audio, and video. The terminology is evolving with the technology, and LLMs are now often called *foundation models* [Bommasani *et al.*] or *frontier models*.

[3] In 1961, Turing laureate Doug Englebart took the contrarian approach of augmenting human intellect [Englebart], which is the term Brynjolfsson used in his paper. We instead use the phrase "improving human productivity" because we think it is easier for the public and policymakers to understand the implications of productivity gains than of augmentation.

[4] This abbreviated history of AI is simplified. In the 1950s there was a fervent energy around the concept of intelligent machines inspired by human brains/intelligence, and during the 1960s the various traditions grew apart. In the big three CS AI departments of the time that were funded by DARPA (MIT, Stanford, CMU), the top-down "symbolic AI" tradition took hold. Rule-based systems mentioned above are just one branch of symbolic AI. Neural networks also got a big boost in the mid-1980s, e.g., the research by Turing laureate Yann LeCun on handwriting recognition using MNIST.

The advent of these large frontier models has raised understandable concerns about energy use of AI. Appendix I covers this topic in detail, but a quick summary is that AI systems today account for under a quarter of 1% of global electricity use, a tenth of digital household appliances like TVs. The International Energy Agency considers even a strong projected increased energy consumption by AI for 2030 to be modest relative to other larger trends like continued economic growth, electric cars, and air conditioning.

While we use the broad term AI, the field is fragmented, covering many technologies. Our discussion will primarily focus on generative and predictive AI systems, with a brief discussion of some other aspects of AI where relevant. Examples are AI assistants (e.g., NotebookLM), chatbots (e.g., ChatGPT), retrieval-augmented generation (RAG) systems[5] (e.g., Perplexity), and generative AI systems (e.g., Midjourney).

## Artificial General Intelligence (AGI)

Before we can get to the impact of near term AI, we first need to consider the prospect of *artificial general intelligence* (*AGI*). An AI system can easily write a new bedtime story daily featuring your children as main characters. A different AI system could beat any human being at the classic strategy game of Go. As of now, no single AI system can do both of these things. Each can deliver amazing capabilities, but they are practically useless if they stray outside their lanes. In contrast to existing AI, proponents argue that an AGI that would be multitalented—capable enough to win strategy games, diagnose diseases, analyze poetry, and contribute to applied computer science innovations that can further enhance the capacity of AGI systems.

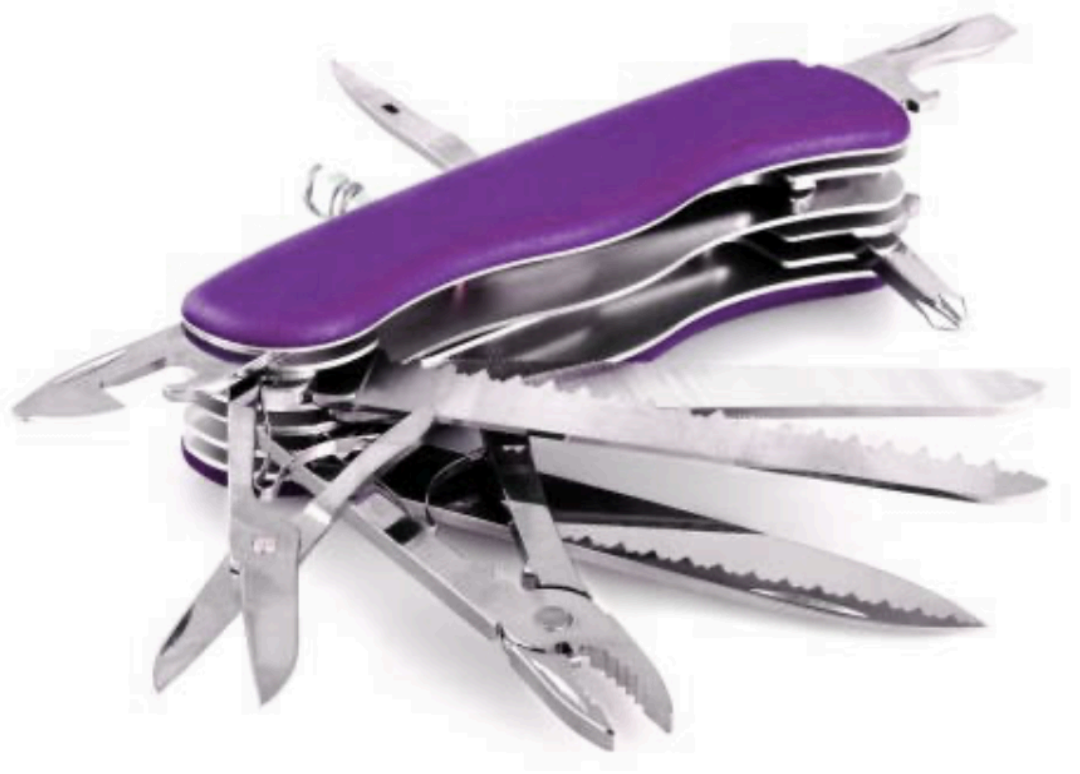

**A general knife.**

AGI has many definitions, but one framework gaining popularity emphasizes *the range of tasks* that an AI system reaches a target threshold compared to people and *how well it compares to human-level performance* for a given task [Morris *et al*.]. Thresholds are labeled based on the portion of people that the system outperforms: competent (>50%), expert (>90%), virtuoso (>99%), and superhuman (>100%). AlphaGo is rated superhuman, but only for playing Go, and is not competent at anything else. This breadth versus depth metric helps clarify AGI discussions.

Tremendous attention is being paid to AGI, deservedly so given its large potential positive and negative impact on the world. We applaud the serious investigations of AGI, including scientific work that aims to clarify relevant definitions and likely impacts.

As we focus on impacts of current and near-term AI systems, we will not discuss AGI further, beyond mentioning that progress on the topic may accelerate both the benefits and risks we outline here.

The next part of the paper delves into the impact of AI systems in the half dozen fields we investigated.

---

[5] *Retrieval-augmented generation (RAG)* is an AI framework that combines LLMs with traditional information retrieval systems to produce more accurate and relevant text.

# II. Demystifying the Potential Impact of AI

## Employment

Our first topic for nearer-term AI is a major concern: the impact on jobs [National Academies]. Indeed, a Global Public Opinion Poll on AI found that the majority expect to be replaced at work by an AI system in the coming decade [Loewen *et al*.].

Technological advancements have long led to the decline of some jobs and the creation of new ones. For the U.S. workforce, 63% had jobs in 2018 that did not exist in 1940 [Autor 2022]. Figure 1 shows examples of four jobs where numbers changed strikingly from 1970 to 2020.

Despite the downside of job disruption, a healthy economy relies on improving worker productivity. Two-thirds of the world's population lives in countries with below-replacement birth levels [Eberstadt] and many nations are facing labor shortages [Duarte]. The U.S. already lacks critical positions as varied as K-12 teachers, passenger airline pilots, physicians, registered nurses, software engineers, and school bus drivers. To supply needed services, high-income countries must either greatly expand their working population or significantly improve worker productivity [Manyika and Spence].

The impact of productivity gains on jobs depends on whether the demand for goods produced by that work is *elastic* or *inelastic*. If demand is *inelastic*, productivity gains means jobs will be lost [Bessen]. For example, agriculture is inelastic in the U.S., so gains meant dramatic declines in absolute numbers (fourfold) and its portion of the workforce (from 40% in 1900 to 20% in 1940, 4% in 1970, and 2% today) [Daly]. If product demand is sufficiently elastic, productivity-enhancing technology will increase industry employment [Bessen].

For example, programmers today are tremendously more productive than they were in 1970—they have more powerful programming languages and tools, plus Moore's Law helped improve hardware a millionfold—yet there were 11 times more programmers in 2020 (Figure 1)[6]. Jet engines and autopilot systems boosted pilot productivity, and even so the U.S. has eight times more commercial airline pilots today than in 1970.

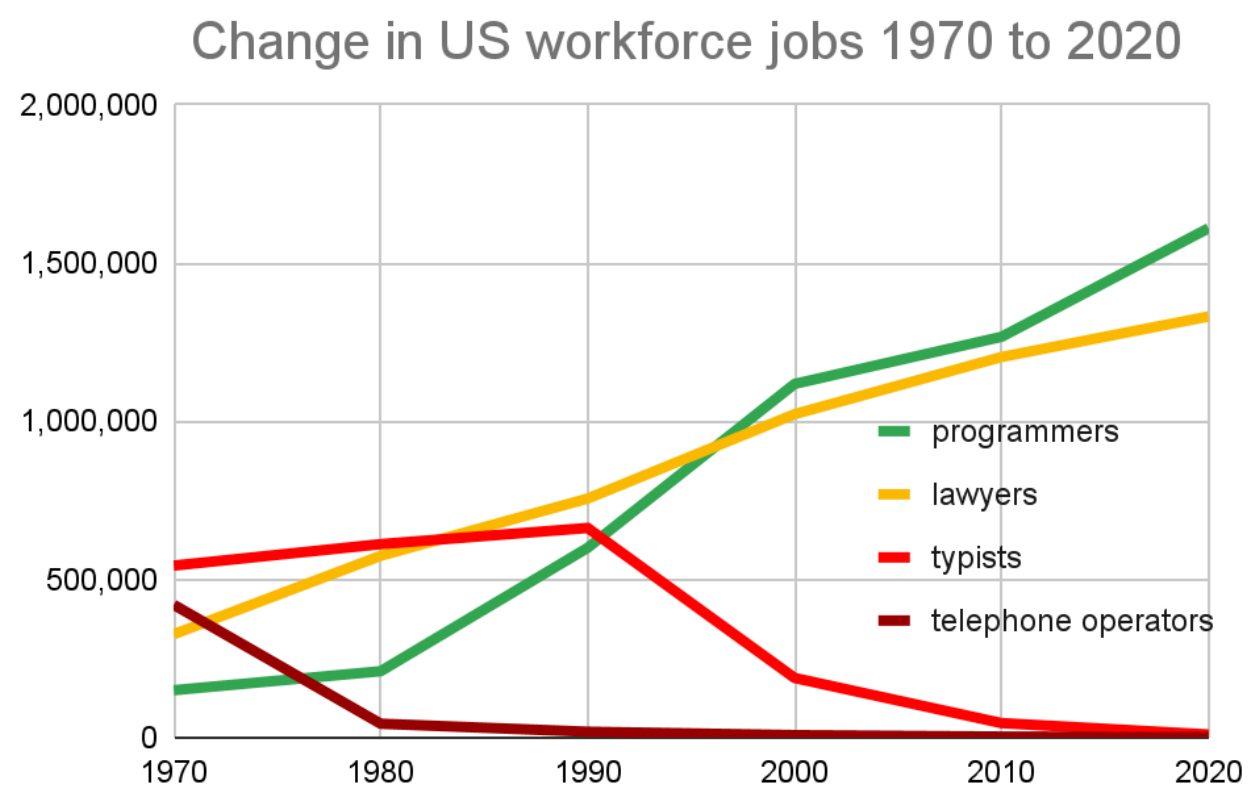

**Figure 1. The number of telephone operators, typists, lawyers, and programmers over 50 years** based on U.S. Census data for each decade since 1970. The number of typists fell ~50 fold and telephone operator jobs shrank ~300 fold, while the number of lawyers grew ~fourfold and programmers shot up ~11 fold. (The U.S. population grew by 60% over that period.) Lawyers now use their computers by themselves, type their own emails and texts, and use their own smartphones. In the law office, a lawyer—plus programmer-developed technology—replaced the typists and telephone operators. In 1970, the ranking by the number of jobs were typists, telephone operators, lawyers, and programmers, which was nearly reversed by 2020. Technology advances were behind the shift in these jobs.

Another perspective on employment is the split between nonphysical tasks and physical tasks. In our view, the main impact of near-term AI systems will be on nonphysical tasks. We think robots will eventually have a large effect on the way in which physical tasks are performed in the world beyond manufacturing, but this may be five or more years behind the use of AI systems for purely digital or knowledge tasks [National Academies]. We don't focus on robotics in this paper, but there has been significant progress over the past decade on robots being able to learn to operate in messy environments and to use a combination of real-world experience, teleoperation, and learning in simulation in order to generalize to new environments and learn new skills.[7] Continued progress could well have large implications for many new areas, including elder care, disaster response, and construction.

[6] A reviewer asked if programmer productivity reduced salaries. The US average salary in 1984 was $20K–$39K, or $46K– $89K in 2016 accounting for inflation. The median salary in 2016 was actually $95K, with the highest 10% making $150K. Both salaries *and* the number of jobs grew.

We don't believe the long-term solution to job disruption by AI systems is government subsidy. The goal should be to continuously ensure that people have the skills to be useful themselves and be appreciated for contributing to society. And even if a concept like universal basic income could scale within the U.S., it's unlikely to scale worldwide.

While the discussion above is about employment, the main issue in the U.S. is not unemployment, but the quality and value of available jobs.[8] While the examples in Figure 1 show lower-wage jobs decreased and the jobs that grew had higher wages, income inequality increased since the 1970s overall in the U.S. because the *average* salary tracked the gains in productivity rather than the *median* salary tracking the gain [Autor *et al*.]. The U.S. has achieved record low unemployment, but the college educated have had much greater economic gains while high school graduates and dropouts got much less, thereby hollowing out the middle class [Autor *et al*.][9].

Our examples so far have been from the U.S., but the impact of AI systems on jobs varies by country, and policies should be tailored accordingly. For example, if we cross borders by traveling 3,000 miles south from Canada (Table 1), AI's imprint alters. Job displacement has a very different impact even with the neighbors of Canada and the U.S.. Unlike Canada, U.S. healthcare is tied to having a job, unemployment insurance is shorter (≤26 weeks vs. ≤45 weeks in Canada), and the federal minimum wage is lower ($7.25 vs. $11.60 in U.S. dollars). Thus, it is much easier to transition to new well-paying jobs in Canada than in the U.S. [Autor *et al*.].

| | Canada | US | Mexico | Honduras | World |
|---|---|---|---|---|---|
| Population (M) | 41 | 335 | 128 | 11 | 8113 |
| Avg. per capita income | $56K | $65K | $11K | $3K | $13K |
| National minimum wage/hour (US$) | $11.60 | $7.25 | $1.80 | $2.12 | n.a. |
| % live on ≤$3.65/day | 0.5% | 0.5% | 10% | 26% | 23% |
| % with college degree | 63% | 50% | 21% | 10% | 8% |
| % no high school degree | 7% | 8% | 56% | 70% | 42% |
| Physicians/1000 people | 2.5 | 3.6 | 2.4 | 0.5 | 1.7 |
| Lawyers/1000 people | 2.5 | 4.0 | 3.4 | 1.7 | n.a. |

**Table 1**. **Population, income, education, and expert availability for four nearby countries in North America and World averages for 2022.** College degree means any degree, including associate, not only bachelor's degree.

While industrialized nations worry about AI systems displacing highly trained workers like lawyers, doctors, and programmers, the low- and middle-income countries face a shortage of such highly skilled workers (Table 1). Making such expertise more widely available in those regions via AI systems could enhance the quality of life *and* accelerate economic growth [Lewis]. Improvements to local economies and critical services may even provide alternatives to emigration for some in middle-income countries [Clemens].

How might AI help? To enhance the quality and value of jobs, AI should focus on empowering humans to do more [Brynjolfsson] [National Academies]. Here are case studies where AI increased productivity, especially for less-experienced workers:

- A University of Minnesota study observed that young attorneys could perform a variety of standard tasks on average 22% faster with AI

[7] Manufacturing robots are currently a $55B industry.

[8] The White House Council of Economic Advisers said it found "little evidence that AI will negatively impact overall employment." A recent report [National Academies] states "the industrialized world is currently awash in jobs."

[9] In the race between education and technology, computerization eroded employment that many noncollege workers did before 1980, increasing inequality by shunting them into low-wage service jobs [National Academies].

systems [Choi].[10]

- A Harvard Business School study found that low-skilled consultants gained more from AI systems than high-skilled consultants (43% faster versus 17% faster), even outperforming those without AI [Dell *et al*.].
- An MIT study documented a 47% speedup for midlevel professional writing tasks with AI systems [Noy and Zhang].
- A Microsoft study showed using AI systems accelerated programming tasks by 56% [Peng *et al*.].

Microeconomic studies do not always lead to macroeconomic results, but early indicators are promising.

While some tasks that humans currently perform will be able to be completely automated (enabling new sorts of tasks and jobs to be tackled), augmenting others will lead to better outcomes, or outcomes that couldn't be achieved with AI systems or with humans alone. Using AI to improve human productivity may also address some challenges —such as safety or misinformation—by the human correcting the suggestions from AI systems when it is mistaken or stepping in when circumstances arise for which the AI system was not trained.

Nevertheless, some AI systems will likely disrupt jobs in fields with inelastic demands for their services. When people are laid off, they will face a range of challenges, including how job losses affect core aspects of individuals' identity, reduce income, and raise questions about a new career to pursue. Moreover, retraining adults for new jobs has had a poor track record.

**AI Milestone: Rapid Upskilling**. To give concrete goals for improving AI's impact, in each section we propose specific milestones. Many current workforce development programs have a modest impact on wages. Our first milestone is an AI system that could reduce the time for workers who are unemployed or in low-income jobs to gain an in-demand skill that is a ticket to the middle class. In the U.S., for example, a goal might be to empower a worker making ≤$30,000 annually to gain a skill that increases their income by ≥$15,000, and to do so in three to six months. (See Appendix II for more details.)

*To give concrete goals for improving AI's impact, in each section we propose specific milestones.*

**AI Milestone: Job Forecaster**. To help workers design their career roadmaps, we need to observe and communicate workforce changes. A *Job Forecaster* would aid workers displaced by AI systems to retrain for well-compensated jobs with growth potential. A real-time Job Forecaster could guide them to an educational path to retrain for promising jobs with future growth [National Academies].[11]

# Education

The next topic is education, where productivity increases and greater equity have long been computing targets. AI systems are affecting classrooms, as 30% of U.S. K-12 teachers and 40% of their students already have used it [Manyika 2024]. Some predict a significant impact from AI on all levels of education [National Academies].

From an employment perspective, we believe that education is elastic, as there is a huge demand for improving the effectiveness and efficiency of learning. Indeed, the U.S. and many other high-income nations face a shortage of K-12 teachers, as

[10] Brazil's supreme court deployed scores of AI programs due to a clogged backlog of 80 million cases. All AI material must be reviewed by a human judge [Guthrie].

[11] All data needed for a Job Forecaster is in plain sight in the U.S.: LinkedIn has many resumes, Indeed.com has the demand for jobs (including job descriptions and number of job requests), and ADP knows the salary per job as it issues almost all paychecks [National Academies]. Information for a Job Forecaster may be easier to collect in countries that have more centralized governance. One technical concern might be that preserving privacy while combining data sets from different organizations and drawing conclusions from the result is too hard, but that technical challenge is well understood today in other contexts. Recent hardware security features (called *enclaves*) enable confidential computing companies like Opaque to provide off-the-shelf security solutions for this situation.

well as STEM graduates who could teach those topics in K-12 schools.[12]

Today's AI tutors such as [CK-12](#) and [Khanmigo](#) likely already help some students. A major educational challenge in the U.S. is that K-12 students in high-poverty schools do much worse on standardized tests compared to students in other countries or to U.S. students from low-poverty schools. Selective use of AI tools might track socioeconomic status, which inadvertently could expand the educational gap between students at high-poverty schools versus low-poverty schools.

Before many schools will deploy AI tools for all their students, they likely first need careful evaluation including *randomized control trials* (*RCTs*) to establish in what circumstances they help or hurt and, if so, by how much. At least in the U.S., it will be difficult to get access to a sufficient number of K-12 students with the desired heterogeneity, their results on exams, and details of their backgrounds, schools, and teachers to assess AI tools properly. A separate challenge would be to convince parents and teachers that their students should be the subject of educational experiments where potentially valuable opportunities would be unavailable to other control groups. Another obstacle to deploying AI tools in K-12 schools in the U.S. is that there are many people beyond the teachers involved in the decision about what tools to use: administrators, school boards, parents, students, and so on. It may be hard to collect the scientific evidence to justify requiring their use, at least in the U.S.[13]

Colleges may offer an easier initial target for assessing the benefits of AI systems in education, as the content is more up to the instructor and their courses are much larger[14]. If we can first demonstrate the success of educational innovation via AI systems for freshman at community and four year colleges[15], it may also simplify the task of deploying it in high schools[16]. Additionally, community colleges play a major role in adult education, including retraining [Schwartz and Lipson]. If AI systems could improve education for retraining, it might partially compensate for the downside of job disruption from AI deployment in inelastic fields.

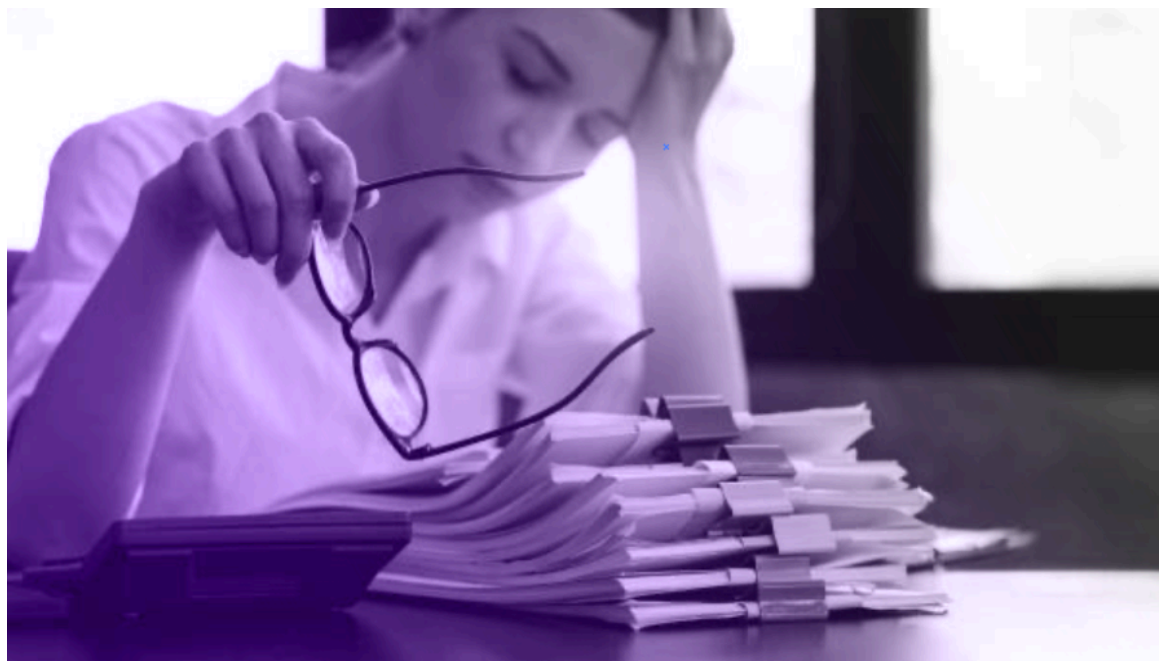

**An AI opportunity.**

However, if some K-12 school districts in other countries or in some adventuresome school districts in the U.S. decide to deploy AI systems for students without waiting for RCT data, one can still collect scientific evidence as a *natural experiment*. The theory is that individuals are naturally exposed to sufficiently different conditions instead of by a researcher's design. Statisticians or econometricians then try afterwards to find many demographically matched groups, and draw inferences. This approach essentially acts as if random assignment occurred, allowing researchers to observe and analyze the effects without actively manipulating variables.[17] These studies usually occur in healthcare where they try to establish causality to convince the government or insurance companies that specific interventions improve both patient outcomes and costs.

---

[12] More than 40% of students in the U.S., Canada, and other OECD nations are at schools affected by teacher shortage ["Class struggle", *Economist*, 7/13/24].

[13] A recent paper claims to be the first to do RCT in K-12 schools [Wang]. They used an AI system to offer actionable guidance for human tutors to give to students. The overall gain was 4%, growing to 9% for lower-rated tutors.

[14] For example, Study Hall is a collaboration between Arizona State University and YouTube.

[15] For example, an RCT showed that Harvard University students taking a physics class learned more than twice as much in less time using an AI tutor than against an active learning model, which already outperforms the classical lecture model [Kestin *et al.*].

[16] Given that high school is mandatory and college is optional, techniques may not transfer despite the similarity of ages of the students.

[17] Such studies predicate involvement of AI in education.

For an AI system to eventually help most students, we believe it must first improve the life of teachers,[18] as teachers largely determine what technology gets deployed. Examples of reducing drudgery might be offering help with lesson plans, progress reports, homework assignments, and grading. To succeed, solutions must be driven by the real, day-to-day challenges teachers face, and they must be aligned with teachers' and students' needs rather than based on the opinions of school boards or administrators.[19]

**AI Milestone: Teacher's Aide**.[20] This milestone's goal is to reduce the unattractive aspects of a teacher's workload, thus improving teachers' quality of life. If AI tools could free up more time for teachers to spend with students and reduce teacher burnout, that by itself would be a major contribution. Metrics of evaluation of a teacher's aide might be hours saved per week and teacher satisfaction/happiness.

**AI Milestone: Empirical Education Platform**. The deployment of teaching tools would be easier if education could be more of an empirical science where RCT tests in heterogeneous environments could evaluate education initiatives. The ideal system would understand the abilities/resources of each student, their classmates, their teachers, their schools, and so forth. An education platform that tries to help students would need to include homework assignments and see results of grading exams.

[18] Google Translate is a huge help to teachers with many students for whom English is a second language.

[19] Kids would prefer to play or do some other form of entertainment, hence laws to force school attendance. In the past, schools even had truant officers to retrieve wayward students. IBM learned this truth the hard way. It has a failed $100M project to build a tutor based on IBM's so-called Jeopardy AI technology because many students wouldn't use it. As Salman Khan explained to us, even if we offered the world's best human tutors to visit students' homes for free, perhaps only 15% of homes would welcome them inside. We need teachers to inspire and enforce use.

[20] Rather than think of AI systems as separate aides to professionals, another perspective is that people would use a variety of AI agents that operate wherever we do work, such as the web, mobile, and real world.

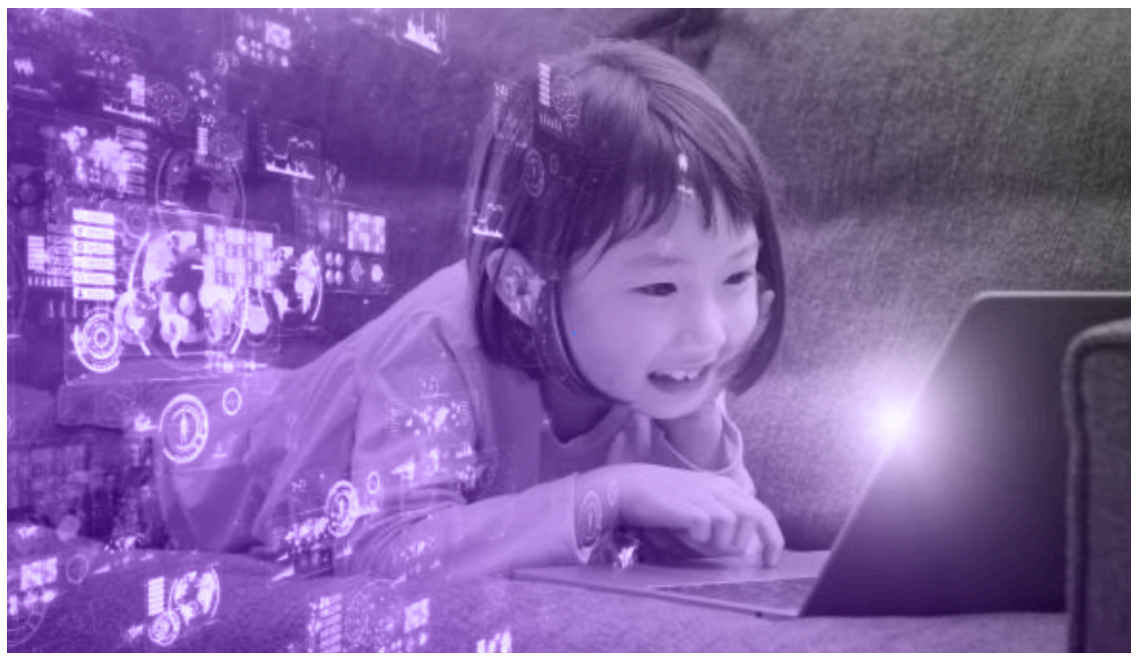

**An AI tutor for all children.**

**AI Milestone: Worldwide Tutor**. Leverage the ubiquity of smartphones[21] by creating a tutor learning tool to accelerate general education for every child in their language, in their culture, and in their best learning style.[22] The focus of this tutor is to help teachers with the challenge of supporting a range of student capability: keeping the high-achieving students engaged while supporting those who are struggling. An advantage of aiming AI tools at teachers is that we can more likely rely on marketplace evaluation and surveys by users rather than RCT assessments of their students, since the goal of using AI tutors in this scenario is to help the *teacher* versus evaluating the benefit to each *student*.[23]

[21] Smartphones are increasingly powerful with their own AI accelerators. Nearly all U.S. teens (95%) have access to a smartphone [Associated Press]. With 7.8 billion people on the planet, there are approximately 6.8 billion smartphones in the world, increasing by 5% per year [Howarth]. AI tools that run standalone on smartphones would expand AI impact to poorer students in high-income countries and to low- and middle-income countries (see Table 1). An example is Read Along, a smartphone-based application in a dozen languages that teaches children to read.

[22] One challenge for students using LLMs is ensuring that AI is not "hallucinating" when it provides answers to student questions. (Hallucination here is the technical term for when an AI system invents incorrect or misleading information.) One potential help is to tie LLMs to answering questions based on textbooks and related authoritative sources. Fortunately, there is a large database of publicly available online textbooks. Such a resource may help instructors to automatically create lesson plans and exams according to educator preferences as well as offering guardrails for questions and answers.

[23] A sign of the potential of AI systems in education in developing countries comes from Rising Academies, a

## Healthcare

The next topic is healthcare, responsible for 16% of the U.S. GDP [Vankar]. Like education, many believe that society should offer high-quality healthcare regardless of the wealth of individuals [Einav and Finkelstein].

We believe that healthcare is also elastic: demand for healthcare will increase more than proportionately as the cost and quality of provided healthcare improves [Baumol]. Indeed, the U.S. and many other countries are facing a shortage of healthcare professionals. Beyond improving the employment prospects of healthcare workers via productivity gains, these tools must also keep healthcare professionals in the decision path for actual recommendations for patient therapy, as AI systems are not guaranteed to make the best recommendation 100% of the time. Since people and AI systems tend to make different mistakes, the collaboration of experts with AI systems might help the quality of healthcare.[24]

Healthcare decisions are made with life-or-death stakes on short timeframes based on complex data, requiring years of specialized training for the best human clinicians. But even then, human specialists have limitations: they are experts in only narrow subdomains; informed by firsthand experience with only tens of thousands of patients; bound by the biases and imperfections of past medical knowledge; available only to the best-resourced healthcare systems; and unable to extract every pattern from the vast sea of medical data.

There are billions of genetic variants, terabytes of medical images, years of lab results, and nontraditional clinical data sources like smartwatch readings, nutritional logs, and environmental risk factors—the complexity of this information inherently exceeds human understanding. Perhaps this is why about 15% of all diagnoses in the U.S. are incorrect [Graber] and why most Americans will be misdiagnosed at least once in their lifetimes. Such errors contribute to ~10% of all U.S. deaths [Ball, Miller, Balogh].

AI has the potential to reduce misdiagnosis rates and evaluate patients more accurately, but AI must first earn clinicians' trust. One way is to first deploy AI in healthcare domains with lower stakes than direct patient diagnosis, such as automating insurance paperwork or transcribing physician notes.

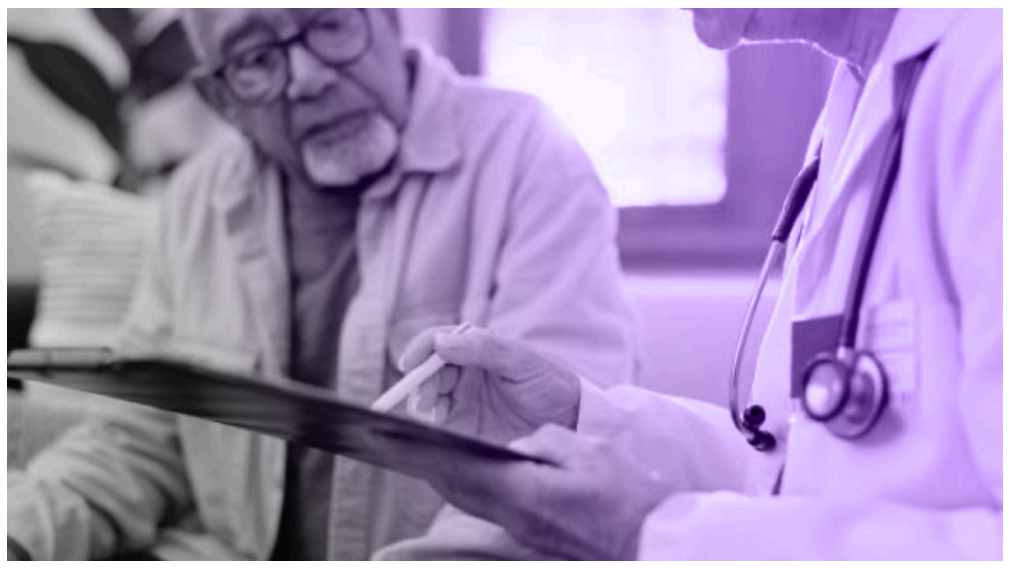

**An AI system could free up more time for doctors to see patients.**

Moreover, AI models can be deployed far more widely than highly trained specialists can: smartphones could put high-quality health expertise in the palm of every healthcare professional. We envision a world where all relevant health-related data and every past healthcare decision can be used to inform every future healthcare decision and benefit everyone.

We are currently far from that world, in part because healthcare professionals are the gatekeepers for deploying AI. AI practitioners should be humble about the enormous challenges that must be overcome to provide advanced tools, especially given unrealistic past claims that AI will soon obviate clinicians. Some of these challenges involve open research questions that arise in the deployment of real-world systems, around equity, usability, robustness, and interpretability, among others.

network of African schools. Rigorously evaluated by Oxford University [Henkel *et al.*], it improves student learning outcomes by a grade level relative to students without it.

[24] One study of an AI system for skin cancer found that the combination of AI and clinicians increased correct cancer diagnosis from 81% to 86% and the accuracy of cancer-free diagnosis. The AI system was more helpful to less-experienced clinicians. In contrast, another study found LLMs did better on their own, suggesting advances needed in the human-AI interface (see Information).

Another important barrier is infrastructure and regulation: few health systems have the infrastructure to easily deploy, update, and monitor algorithms, and health systems are cautious in light of strict healthcare regulations.

Though the path is long between the current state of medical AI and the world we envision, the rapid pace of progress in developing the underlying technology is cause for optimism. However, progress relies critically on data availability [Mullainathan and Obermeyer]—large, diverse, multisite cohorts to ensure models perform robustly and equitably across many populations and conditions, as well as techniques like federated learning, which allows AI models to be trained on many distinct pools of data without centralizing any of the raw data.

Policymakers and stakeholders should permit and encourage healthcare organizations to participate in multi-institution collaborations to use de-identified data to train machine learning models for the benefit of their patients and others around the world. Policymakers and stakeholders should also insist on open standards for the interchange of health data and for the inclusion of AI-based predictions and guidance in clinical workflows.

In light of these challenges, we see two complementary directions for fulfilling the potential of health AI if we first succeed at deploying tools that improve the workweek of healthcare professionals. The first is deployment of health AI systems designed for specific, important tasks—e.g., screening patients for diabetic retinopathy [Gulshan *et al*.] or detecting cancer from mammograms [Lång *et al.*]—and collecting convincing proof (e.g., from RCTs) that they improve patient outcomes. Healthcare economists often do the critical studies with proper causal inference established to convince payors (the government and insurance companies) that specific interventions improve both patient outcomes and cost efficiency. Such successful economic studies efforts in the domain of AI and healthcare efforts remain relatively rare and can be enormously impactful. They also provide insight into challenges that arise in real-world deployments and test the political will to overcome the regulatory and institutional barriers to gain access to such data.

Complementing the deployment of *narrow* AI systems—designed for specific tasks—a second promising direction is the development of *broad* medical AI systems [Moor *et al*.] that learn from many data modalities and are designed for many tasks.

**AI Milestone: Healthcare Aide.** Before aiming for tools that help patients directly, AI practitioners should first develop tools to reduce the paperwork and drudgery currently required of healthcare professionals.[25] If AI tools can make their jobs less tedious, they could reduce burnout by healthcare professionals and set the stage for later tools that can help patients more directly. Improving the skill sets and productivity of nurses and physician assistants could also give more patients access to quality healthcare in the many regions around the world facing doctor shortages (see Table 1 above).

**AI Milestone: Narrow Medical AI.** Deploy a health AI system designed for a specific, important task such as predicting patient deterioration in the ICU [Escobar *et al*.] The impact of the milestone could be measured in the number of patients who benefit.

[25] A reviewer proposed a *patient* aide to clarify patient-doctor communications (including with nonnative speakers) to help them explain their symptoms and to help explain a doctor's advice and remind them how to follow it. *Nonadherence* leads to serious health risks. Patients who didn't follow orders were twice as likely to be readmitted to the hospital within 30 days [Mitchell *et al*.]. In 2003, about 20% of U.S. Medicare patients were readmitted to hospitals within 30 days, which cost Medicare ~$17.4B [Jencks *et al*.].

**AI Milestone: Broad Medical AI.** In contrast, a broad medical AI system would learn from many data modalities—images, laboratory results, health records, genomics, medical research—to carry out a diverse set of tasks, and be able to explain its recommendations using written or spoken text and images. Such tasks might include bedside decision support, interacting with patients after they leave the hospital, and automatically drafting radiology reports that describe both abnormalities and relevant normal findings while taking into account the patient's history [Moor *et al.*]. To realize this milestone, we need to define and maintain metrics and benchmarks that would measure progress toward a broad medical AI system.

# Information / News / Social Networking

While education and healthcare offer targets that could potentially increase the upsides of AI, the goal of this section is to reduce risks associated with one of the widely feared downsides of AI. In most of the scenarios discussed so far, AI systems act to provide information, such as personalized tutoring in education or helpful diagnostic information in medicine. However, AI may accidentally generate incorrect information (*misinformation*), or be used to maliciously generate incorrect information (*disinformation*) such as in the case of false news presented as fact (especially an issue in political election tampering), or generated visual imagery or spoken audio presented as real.

A related concern is built-in harmful bias affecting critical decisions like criminal sentencing or mortgage lending. As AI systems become more independent in their interactions with humans and other AI systems, the potential benefits but also the risks posed by misinformation, disinformation, and bias grow, and whether an AI system can be trusted to engage in action aligned with the interests of users (or society) becomes even more important.

To achieve the potential of AI, we must build ways to maximize the benefits of AI-provided information but mitigate the effects of misinformation, disinformation, and bias. While the threats of disinformation to personal well-being and international security are clear, the threat of partial misinformation or bias is more nuanced: AI systems are imperfect, yet people expect their tools to be reliable.

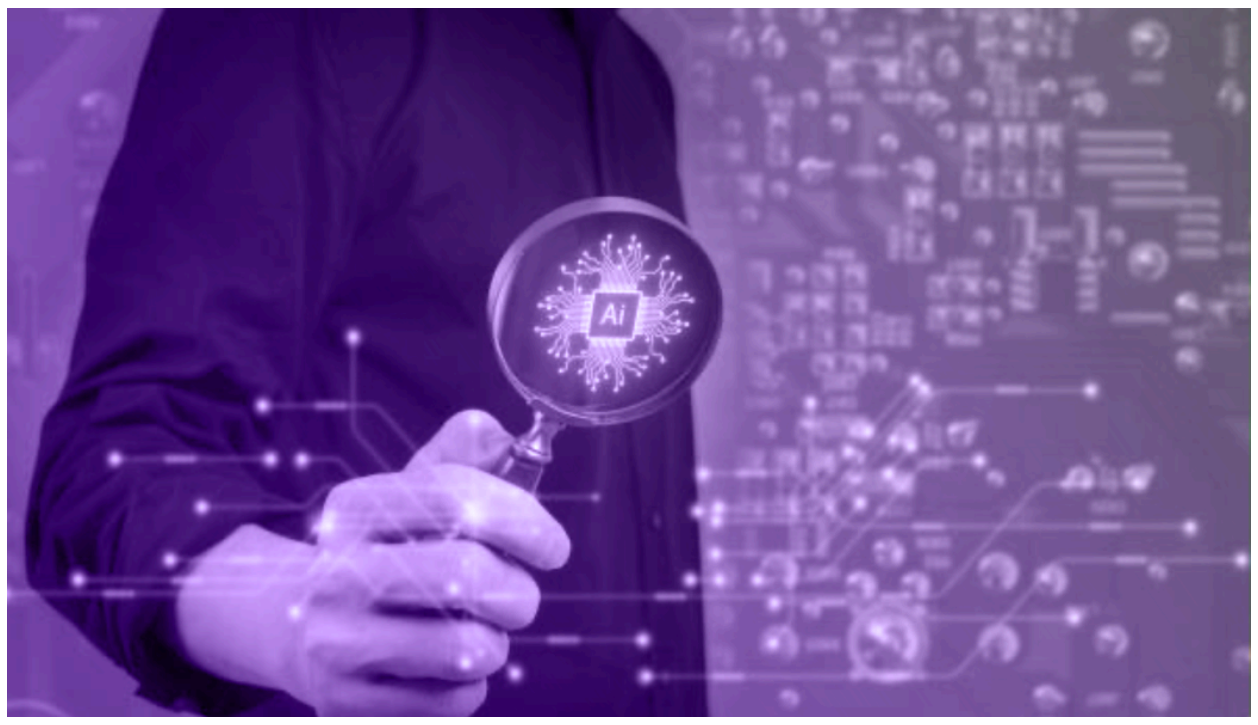

**Some AI systems can detect whether or not text or images were created using their GenAI tools.**

As a result, people tend to overtrust AI systems not only in cases when it is purely wrong (e.g., recommending an unsafe drug dosage) but also only partially correct (e.g., radiology support that finds some tumors but not all) or biased (e.g., favoring some job candidates over others on the basis of protected characteristics). Market forces may add further gray zones with respect to bias if, for example, information providers allow sponsors to pay in exchange for AI-generated answers that favor certain products or viewpoints. (We further discuss ways of mitigating bias in the Governance section below.)

Similar problems will arise as AI systems begin to behave more independently, engaging in transactions or other interactions on behalf of users, with AI systems potentially overtrusting other AI systems. Successful ecosystems of AI systems and human agents will require the ability to assess whether the AI systems have behaved in a trustworthy manner, used relevant information effectively, and managed to navigate through misinformation.

Solutions to overcoming AI misinformation, disinformation, and bias challenges will require not only high-quality AI systems, but also effective human+AI (and AI+AI) interaction—an area that has received significantly less regulatory and research attention. We must develop methods that provide users, developers, and regulators with control over and understanding of AI systems.

We expect these methods will include new types of user interfaces that display useful context along with an AI system's answer, such as the AI system's confidence level, or citations of the source of information included in the answer. At a lower level in the AI toolchain, such understanding will require techniques for interpreting the internal mechanisms of AI models that could be used to know if, for example, a model is inappropriately considering race or gender and generating a biased answer; using reasoning versus referencing something it has memorized; or even if it is lying versus being truthful.

One piece of good news is that AI systems show early evidence of identifying disinformation. OpenAI created a tool for its DALL-E LLM that correctly identified 98% of the images it generated and only misidentified 0.5% of non-AI generated images as ones it created [OpenAI]. Alas, it did not work as well with GenAI programs from other sources. Another recent paper demonstrated how watermarks could be deployed to identify text that is machine-generated text [Dathathri *et al*.].

AI systems also can help with civic discourse. One study compared discussions between people on opposing sides of an issue [Argyle *et al*.]. Conversations in which the AI system would make suggestions on how to rephrase comments and questions more diplomatically before being exchanged led to much greater understanding between the two sides than conversations without the help of AI [Bail *et al.*]. The study concludes:

> *"Though many are rightly worried about the prospect of artificial intelligence being used to spread misinformation or polarize online communications, our findings indicate it may also be useful for promoting respect, understanding, and democratic reciprocity."*

Another study used an AI system to hold discussions with conspiracy theorists. Conspiracists often changed their minds when presented with compelling evidence [Costello *et al*.]. The researchers' summary is:

> *"Conspiratorial rabbit holes may indeed have an exit."*

A recent paper reports on two promising current platforms for civic discourse and proposes required features for a successful platform: The School of Possibilities and Polis [Tsai and Pentland].

**AI Milestone: AI-mediated Platform for Civic Discourse.** This milestone is for a platform that mediates conversations or attitudes to enhance public understanding and civic discourse. Features measured could include the breadth of the topics covered, the effectiveness of the tool versus the difficulty of each challenge, and so on.

**AI Milestone: Disinformation Detective Agency.** If each service could create a highly accurate detection tool for its GenAI software, then in theory we could, with high probability, identify many deepfakes. One problem is that if the tools were made publicly available, it would likely lead to an escalation battle between the deceivers and the detectors, as deceivers could use the tools repeatedly to revise their deepfakes until they were undetectable. Taking a page from the doping problem in athletics, we might reduce its use if international organizations held such detection tools, and in response to requests from political watchdog groups or law enforcement agencies could render judgments on probabilities of current or past[26] images being deepfakes.[27,28]

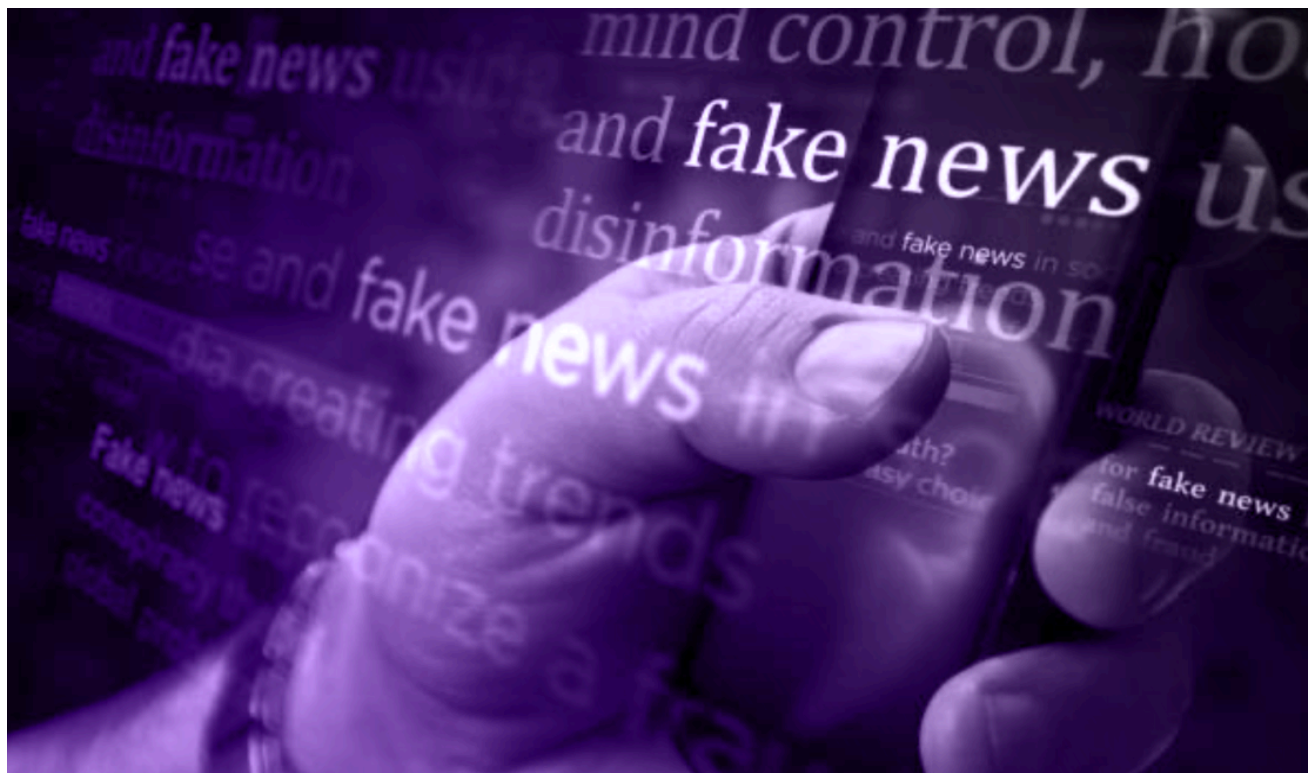

**AI can find exits from conspiratorial rabbit holes.**

**AI Milestone: Controllable AI for Curating Information Consumption.** The goal is a personalized, controllable agent for information consumption that balances the values of personal preferences, learning, and exposure to new perspectives. This AI agent can focus on information curation problems in the domains of interest, i.e., information, news, and social media. Notably, the milestone will require an agent that goes beyond fact-checking into curating a controllable information diet, i.e., the agent can be highly personalized via both implicit and stated preferences and balance instantaneous preferences versus ideal and societal values (which measure long-term positive impacts).[29] The solution could involve a single, steerable AI system or a framework to transform a model into an effective form of the goal.

## Media / Entertainment

Unlike education and healthcare, many areas of entertainment are inelastic [Pannell]. It is not obvious that if AI improved the productivity of fine artists and graphic designers that the market would grow to accommodate many more paintings and designs. And there is no shortage of fiction novelists; publishers have inboxes full of unsolicited manuscripts. Even a successful author like Stephen King had to adopt a pen name because his publishers feared he would write more books than his market would bear.[30]

Journalism is very different from writing fiction. Journalists must write to a tight deadline while simultaneously being under tremendous pressure to not make mistakes in their reporting. While there are some tedious news chores that are better left to AI systems—turning quarterly financial reports from companies into text or reporting on high school sports—investigative journalism and many other tasks are not low-hanging fruit for AI. CNET secretly (and now famously) tried using AI to write dozens of feature stories, but then had to write long correction notices about "very dumb errors."

We should develop AI methods specifically for fact-checking to enable a journalist AI tool that reliably and quickly copyedits reporters' stories, including checking for spelling of names and places. Given the importance of journalism in a democracy despite its current precarious economic state, AI tools that reduce the stress and burnout of journalists are likely a significant contribution to civil discourse.

The impact of AI systems on the movie industry is much harder to predict. The special effects using toy models and stop-action films before the 1980s have been replaced by computer-generated images, which surely employ many more people even if the skill sets are very different.[31]

[26] Doping eventually led to international organizations taking blood samples from athletes to test immediately for illicit substances. A further hindrance is to keep the samples for several years, so that future tests could detect past cheating to discourage bad behavior even if it were undetectable today.

[27] If the GenAI tools included watermarks in their images, they could point to the customers who created them. One recent paper demonstrated the watermarks could even be deployed to identify text that is machine generated [Dathathri *et al.*].

[28] Laws have already been passed to punish the use of disinformation in some circumstances. For example, California passed a law barring publications of deepfakes about candidates 60 days before an election. More than 20 other states have passed a related law or are awaiting signatures of their governors.

[29] Some early work in social media shows how optimizing for ideal values can improve long-term satisfaction [Bernstein, *et al.*] [Khambatta *et al.*]

[30] Over his 50-year career, so far he has written seven novels as "Richard Bachman" and 58 as Stephen King.

Neal Stephenson gives the analogy about the impact of AI systems on entertainment by trying to explain the impact of movies to stage actors in 1900. Back then they memorized plays, performed every night in front of a live audience with other actors, and needed to project their voices to reach the back of the theaters. Imagine if these actors were told that the future includes:

- individual performances in a warehouse
- no live audience
- no need to project their voices
- sometimes no other actors
- A single performance recorded and repeated in thousands of theaters for months.

They would probably fear for their future and for the future of acting as a profession. Instead, live theater is still healthy on Broadway and the West End alongside cinema because audiences get different experiences from these varied performances.

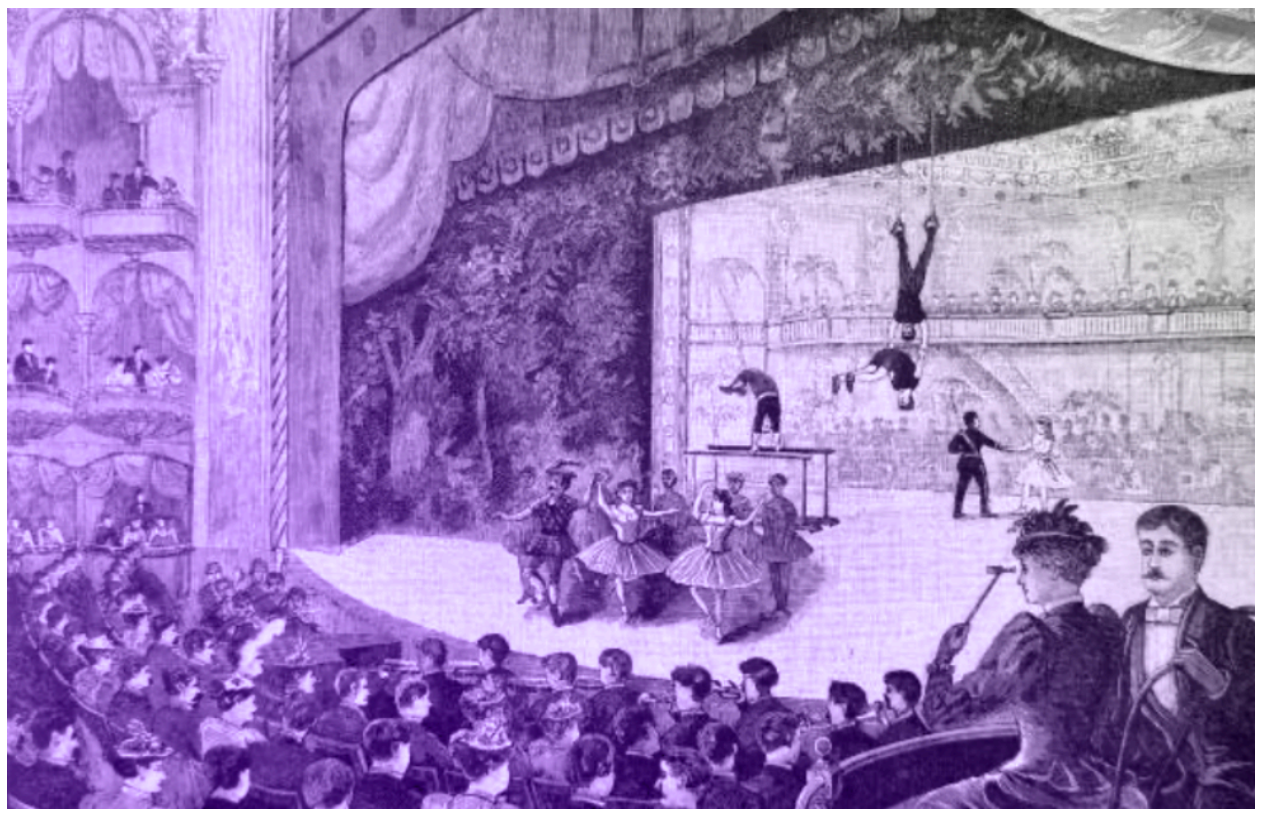

**Entertainment before movies. Live theater on Broadway is still healthy alongside cinema.**

Today's image and video tools allow amazing control over every pixel on the screen, but they are infamous in their difficulty to use given the need to set hundreds of virtual knobs. If the tools themselves make it much easier for the director to keep control of the thousands of microdecisions need to create art [Stephenson]—or if an AI aide can step in and remove the drudgery of such tools—then we can imagine a Cambrian explosion of feature-length films where the movie studios are no longer the gatekeepers of what can be made since filmmakers would no longer need to raise millions of dollars beforehand. If we think of entertainment as a storytelling industry, one outcome is that AI systems could help more individuals to tell more stories.

If such advances enable a thousand future Martin Scorseses or Steven Spielbergs to make movies with tens of their friends, it's not clear if the movie industry will be smaller, even if the job makeup for portions of the industry would be very different. The future of cinema could be more like what is happening today with television and video sharing platforms today for younger audiences. One report found that Gen Z watched only 20 minutes of live TV daily but more than 90 minutes on platforms like TikTok and YouTube.

**AI Milestone: A Journalist's Aide.** Rather than trying to write stories, this tool would quickly check for mistakes in news drafts based on the information available on the internet and from other AI tools. Instead of automatically correcting the text, like a spell-check or grammar checker, it would highlight places that it couldn't verify, potentially showing the conflicting sources to warn the journalists. While journalism might not be a large market commercially, open-source development has long led to useful, production-quality, free tools in computing. It's easy to imagine an important, worthwhile mission like supporting journalism inspiring such an effort.

**AI Milestone: Copyright Detector/Revenue Sharer.** For creators of original artwork, rights/usage/remuneration issues are an area of immediate concern. Ideally, when use of copyrighted work was detected, it would lead to distributing funds accordingly to the owners of that work. A revenue-sharing component along with the ability to detect unlicensed use are important factors. A related target would be a way to reward those who develop licensable data sets specifically designed to improve the quality of AI training in specific narrow domains.

[31] The CGI market was $3.2B in 2020, growing 20%/year.

# Governance / National Security / Open Source

As AI systems permeate workplaces and daily life, policymakers and the public will face a range of choices—some relatively familiar and others opening up cans of worms of potential policy and regulatory challenges. Like aviation, television, and the internet before it, AI promises both bountiful rewards and potential perils. Yet its nature as a rapidly evolving, general-purpose technology that can be used to outsource human decision-making sets it apart. As people come to use AI systems more often in their workplaces and daily lives, policymakers are confronted with decisions not only about whether or how to design new policies for particular AI systems or uses, but also how to interpret existing laws that already govern what people do.

Existing laws already govern many AI applications. Tort law, for instance, holds entities liable for unreasonable risks when deploying AI systems in services like accounting. Sector-specific regulations, such as FDA oversight of AI-enabled medical devices or provisions of international humanitarian law governing certain military decisions, remain applicable. The challenge lies in interpreting these laws for novel AI use cases, an endeavor that will often demand fact-specific judgments and greater knowledge from policymakers about the technical attributes of AI systems.

Another challenge is to address certain limited gaps in existing law with carefully targeted policies taking into account the unique capabilities and benefits of advanced AI systems, and deploying approaches that recognize how quickly the technology is changing. Safety and security testing for AI models—including appropriate backups and fail-safes—for managing critical infrastructure, such as power grids or air traffic control, is crucial. Countries need strategies to determine how the most advanced AI models enable adversaries to engage in cyberattacks or to design specialized weapons, and to reduce those risks. Nations vying for AI supremacy must balance cooperation between government and private companies with antitrust concerns.

The debate over whether to open-source AI models exemplifies the nuanced approach required. While sharing model weights and technical details can spur innovation, it may also aid adversaries. The devil is in the details: legal terms of sharing, built-in safeguards, and the extent of disclosed information all factor into the equation. Accordingly, policies designed to limit the risks of open-weight model releases must be carefully designed to retain as much as possible the benefits of openness while limiting the ease with which openly available models can be reconfigured for malign use [Bateman *et al*.].

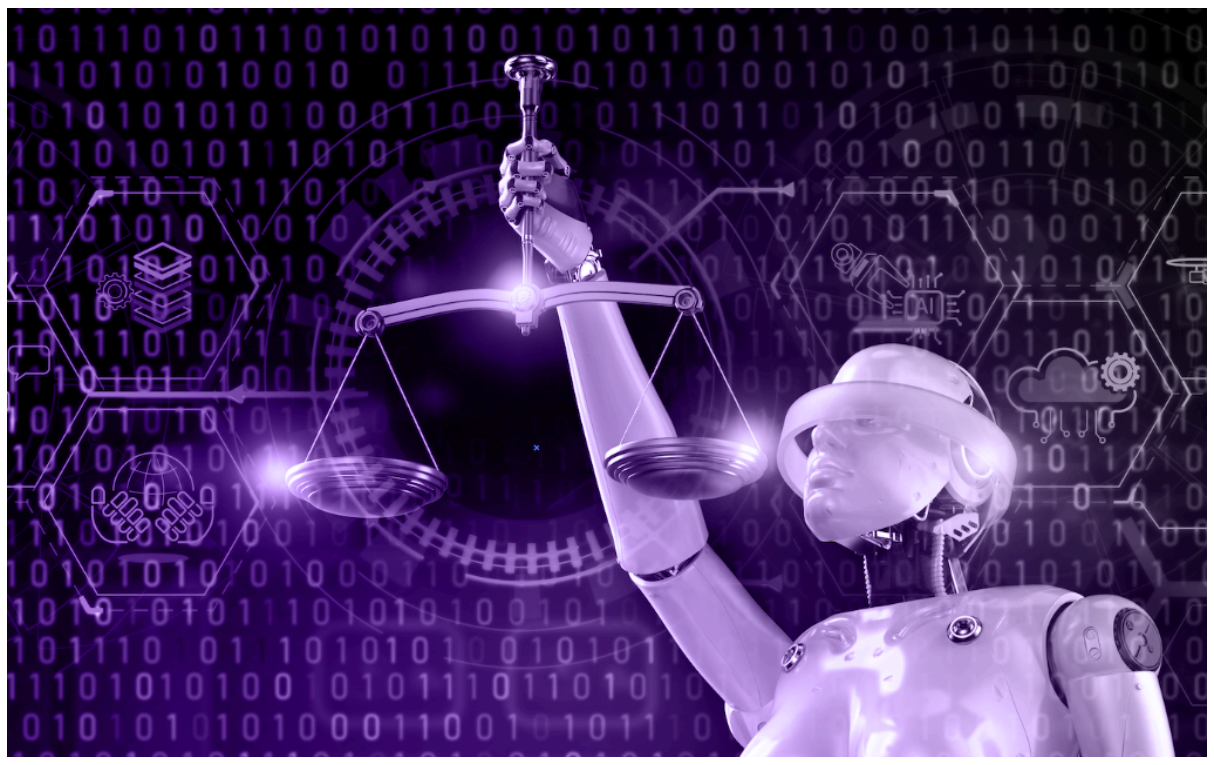

**Existing law can govern many AI scenarios.**

To address the issues above, policymakers should consider seven key principles:

**1. Balance benefits and risks:** Focusing solely on perils can stymie beneficial outcomes and innovation [Ng]. Autonomous vehicles, for example, may drive exceedingly cautiously and slowly in extreme driving conditions but could prove safer and more convenient in normal circumstances. Increasingly capable AI systems can behave in unexpected ways—particularly vast numbers of AI "agents" interacting with other AI systems on behalf of their users—and facilitate some malicious activities. But it can be as big a mistake to ignore the benefits as it is to ignore the risks. We must also bear in mind the difficulties inherent in governing fast-evolving technologies rather than their specific applications, and be wary of false dichotomies [Ng]. For example, by gradually but steadily expanding the range of situations in which autonomous vehicles are permitted, we can get the benefits of safety and convenience of autonomous vehicles in normal circumstances and the expertise of a human driver in more unusual circumstances where the autonomous vehicle would struggle.

**2. Holistic and transparent impact assessment:** Impact assessments of evolving AI systems play a key role in ensuring people, organizations, and governments better understand the potential contributions of AI systems as well as their risks and limitations. AI's effects span democracy, healthcare, education, employment, economic competition, and national security. Policy discussions must encompass this broad spectrum. Because AI systems can have heterogeneous effects, these impact assessments must be as reliable as possible and made broadly available for individuals and advocacy groups so they can alert policymakers of concerns.

---

*The decisions made today will shape the AI landscape of tomorrow, influencing everything from economic competitiveness to social stability. At the dawn of practical AI, thoughtful governance is not just desirable—it is essential.*

---

3. **Leverage existing legal frameworks:** Rather than crafting entirely new regulatory schemes, adapt and apply current rules where possible. Laws regulating fraud and other forms of illegal behavior are applicable whether AI systems are used or not. Many laws apply to AI systems when used in particular contexts or for certain purposes: AI-enabled medical devices are already regulated by the FDA in the U.S., and violations of anti-fraud statutes or securities regulations are subject to punishment whether they are committed with or without AI. Governing day-to-day applications of AI is more about thoughtful application of the rules and standards we already have than about comprehensive new schemes that seek to regulate the use of AI systems in every context. In particular, we should ensure that AI systems are held to at least the same legal standards as human decision-makers or non-AI systems in the same context, so that AI systems cannot be used to avoid responsibility.

**4. Fill in the gaps:** Even if we rely on careful interpretations of existing laws to handle 80% to 90% of the AI policy challenges, carefully designed new policies will help society manage a technology that can turn some forms of intelligence into a manufactured commodity. When the most advanced AI models are deployed to manage critical infrastructure, for example, credible safety and security testing—as well as appropriate backups and failsafes—can help reduce the risk of mishaps. Testing and suitable design changes can also limit the dangers of misuse by people trying to strip off model guardrails so they can perfect more advanced cyber-intrusion exploits. Given that industry is advancing the AI frontier, we will need frameworks for cooperation between government and private companies taking account policy considerations ranging from national security to economic competitiveness, antitrust concerns, and the parameters of AI audits.

**5. Mitigate bias:** There are myriad past examples of biased AI systems, including automated speech-recognition AI systems that performed less accurately for Black speakers than white speakers; image analysis systems that performed less accurately for people with darker skin; and genetic risk scores that performed less accurately for people of non-European ancestry. To mitigate these concerns, regulators should encourage best practices following lessons learned from these past failures.

For example, AI models should be trained on datasets that reflect the diversity of the people they are intended to serve. Evaluations should assess subgroups of the population (e.g., defined by race, gender, and other sensitive attributes) not just the population as a whole to reveal disparities in performance. If predictive models are used, they should be assessed for target variable bias in which the target the model is trained to predict is a biased proxy for its intended target. Removal of sensitive attributes from a model's inputs should not be viewed as a panacea to bias, both because the model can reconstruct them through proxies and because removal of sensitive attributes can sometimes worsen bias. In addition, the potential of AI models to *reduce unfairness* should not be overlooked, since the status quo of human decision-making is often profoundly imperfect, and AI models can serve as useful means for advising human decision-makers and detecting and reducing human bias.

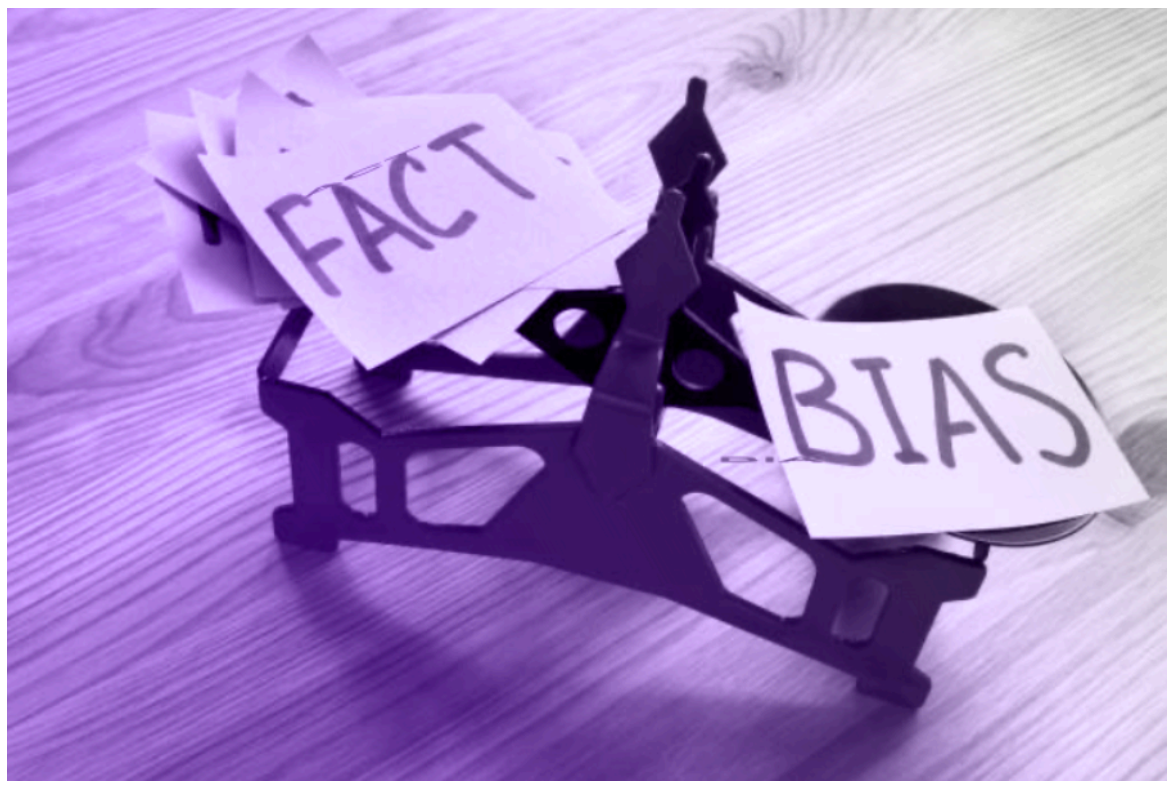

**Can AI overcome past cases of bias to reduce unfairness?**

**6. Invest in public interest and national security:** Some promising policies involve investment and infrastructure rather than regulation. Policies can encourage applications of AI not yet fully embraced or funded by the market or capabilities critical for national security. Companies working at the frontier face energy- and land-use-related challenges that will require coordination with the government. Another example is to support investment in research infrastructure to keep universities engaged in advanced AI research.

**7. Embrace iterative policymaking:** The rapid pace and general purpose nature of AI development demands continuous evaluation and refinement of policies. AI-enabled platforms can facilitate both the evaluation as well as public consultation and deliberation.

The challenges are substantial but not insurmountable. Public agencies need tech-savvy talent and computing resources, and in some cases new procedures, to keep pace with innovation. International cooperation must be balanced against national interests. The opaque nature of some AI systems may clash with requirements for interpretable government decision-making. Yet, with judicious application of existing laws, carefully crafted new policies, and the right personnel and technical resources to enable faster and smarter decisions in the public sector, society can harness AI's potential while mitigating its risks. The key lies in striking a balance between fostering innovation and ensuring responsible development. The decisions made today will shape the AI landscape of tomorrow, influencing everything from economic competitiveness to social stability. At the dawn of practical AI, thoughtful governance is not just desirable—it is essential.

**AI Milestone: Recent Government/Industry Collaborative Successes.** This milestone features collaboration between government agencies and private industry worldwide to develop policies and technology that enhance the upsides of AI and dampen its downsides.

**AI Milestone: Implementable AI Audits.** A number of frameworks have been proposed for audits and impact assessments of AI (e.g., Canada's directive on automated decision-making, WEF's AI procurement in a box, EU AI Act guidelines); However,the procedures for how to actually perform an audit or assessment— the level of expertise of the auditor, the specific tests that must be run, the specific information that must be reported—often remain vague. This milestone is a public-private partnership that establishes these criteria.

**AI Milestone: Equity-improving AI.** Equity goes beyond equality—which is mathematically convenient to formulate (e.g., as equal accuracy rates across groups), and recognizes that individuals have unequal circumstances warranting different needs. Thus, equity is difficult to quantify and operationalize. In large-scale deployments (e.g., for governance), operationalizing equity requires balancing stakeholder preferences and dealing with missing or incomplete information (e.g., due to privacy concerns). We believe the most impactful areas for AI equity interventions are applications in governance, where AI may be deployed for decision-making that affects billions of people and has significant social and economic outcomes. In such settings, this milestone will involve either deploying an equitable AI decision-support system or conducting an analysis that relies on AI, along with clear real-world evidence that it has improved some measure of equity.

## Science

The poster child for AI in science is protein folding. Remarkably, the scientists involved received the Nobel Prize just six years after the first version of AlphaFold was released.[32] It addressed a 50-year-old puzzle: how to predict protein structures from their amino acid sequences. Michael Levitt, who received the Nobel Prize in a related field, said AlphaFold advanced the field by 10 to 20 years. More than 2 million scientists in 190 countries have used it. The Nobel committee assesses the impact as follows:

> *"Among a myriad of scientific applications, researchers can now better understand antibiotic resistance and create images of enzymes that can decompose plastic."*

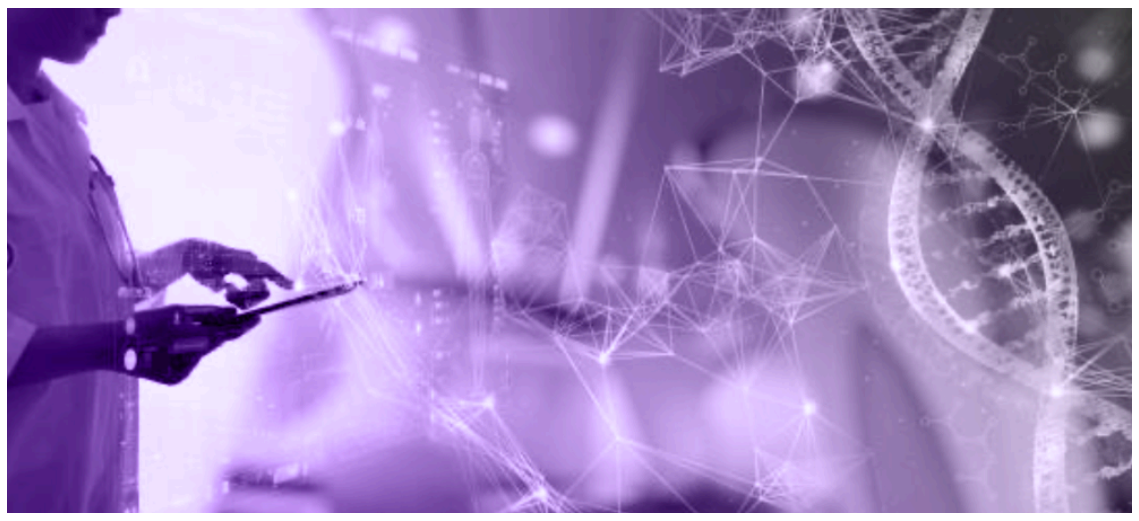

**AI systems for science *and* scientists.**

Nobelist John Jumper told us one of the enabling artifacts was the Protein Data Bank (PDB). Started in 1971, this peer-reviewed repository holds information about the 3D structures of proteins, nucleic acids, and complex assemblies. It holds 200,000 examples and is one of the best databases in biology, which made it an attractive target for AI systems. More curated datasets would create more opportunities for scientific advancement via AI systems, as progress can be limited by the availability of high-quality data. In many science fields—chemistry, materials science, biology—researchers are now using "self-driving labs" that combine robotics and AI systems to reduce the time to make a new scientific discovery.

It was only a dozen years ago that neural networks started outcompeting AI alternatives. It is hard to overestimate the current excitement about the promise of AI within the broader scientific community [Hassabis and Manyika]. Here are more examples:

- **Black hole visualization.** A partnership between AI systems and astrophysicists reveals events that are otherwise unseen. A Caltech research group used generative AI to make 3D videos of black hole M87*—famously the first black hole to appear in an image—and of the flares that occur around the black hole at the center of our galaxy [Lin *et al.*] [Levis *et al.*].
- **Flood forecasting.** Researchers were able to develop an AI model that achieves reliability in predicting extreme river-related events in ungauged watersheds at up to a five-day lead time [Nearing *et al*.] with reliability matching or exceeding those of instantaneous predictions (zero-day lead time). It now covers hundreds of millions of people in over 80 countries.
- **Materials discovery.** The Graph Networks for Materials Exploration (GNoME) generates novel candidate crystals and predicts their stability [Merchant *et al*.]. GNoME successfully discovered 2.2 million new crystals of which 380,000 are the most stable, making them candidates for synthesis. Microsoft's MatterGen is another example that focuses on creating synthesizable materials with specific desired properties, such as chemical, symmetry, or electronic/magnetic characteristics [Zeni *et al*.].

[32] The Chemistry Nobel prize was shared with David Baker, who won for computational protein design.

- **Weather prediction.** AI models run 1,000 to 10,000 times faster than conventional weather models that run on supercomputers, allowing more time to interpret and communicate the predictions as well as reducing cost and energy consumption [Wong]. The AI model GraphCast predicts weather conditions up to 10 days in advance more accurately and much faster than the industry gold-standard weather simulation system [Lam *et al.*]. It can predict the tracks of cyclones with great accuracy further into the future, identify atmospheric rivers associated with flood risk, and predict the onset of extreme temperatures.
- **Contrail reduction.** An AI model identifies areas where airplane contrails are likely to form, allowing for flight rerouting to reduce the climate impact of air travel. Initial results tested in partnership with American Airlines showed a 54% reduction in contrails with minimal fuel increase [Platt]. Reducing the frequency of contrail formation could have a significant impact on emissions from air travel, as they account for ~35% of the global warming impacts of the aviation industry [Geraedts].

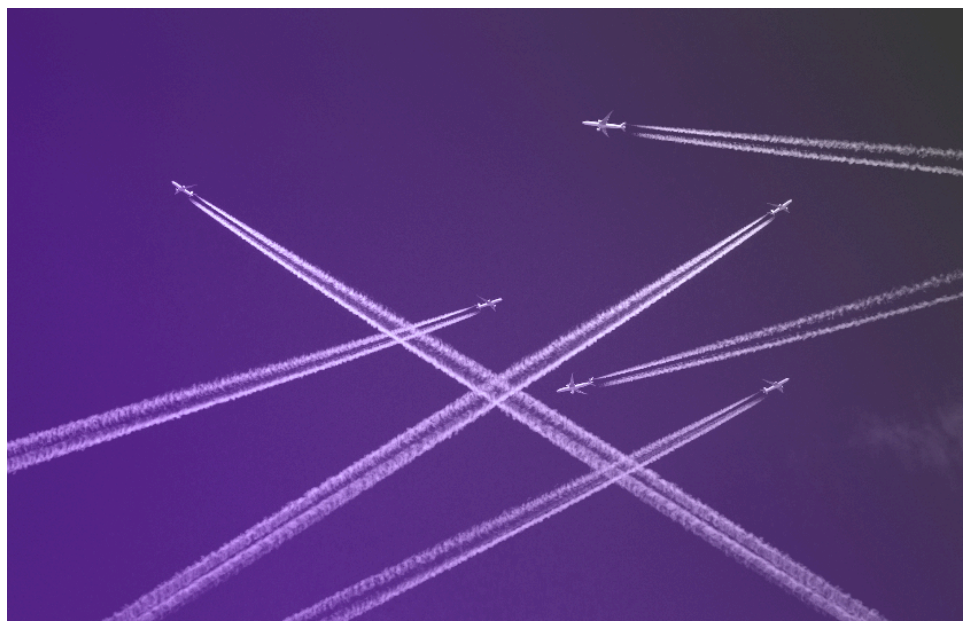

**Reducing contrails reduces climate impact.**

- **Controlling plasma for nuclear fusion.** Researchers used an AI system to autonomously discover how to stabilize and shape the plasma within an operational tokamak fusion reactor. Stabilizing plasma is a critical step on the path toward stable fusion [Degrave *et al.*].

These examples of AI's impact on science have been vetted by peer review and published in prestigious scientific journals, so the research itself is a few years old. More experiments are underway on many topics using the more powerful recent AI models. As one report states [The Royal Society]:

> *"The increased investment, interest, and adoption within academic and industry-led research has led to a 'deep learning revolution' that is transforming the landscape of scientific discovery."*

A few successes could plausibly accelerate progress on some of the United Nations' 17 Sustainable Development Goals (SDGs). This compelling list includes global targets for 2030 on poverty, hunger, healthcare, education, the environment, and other meaningful topics.

Dario Amodei, a founder and CEO of Anthropic, argues that with more powerful AI a system could perform, direct, and improve upon nearly everything biologists do [Amodei].[33] He projects that such an AI system might enable biologists and neuroscientists to make 50 to 100 years of progress in five to 10 years.

Breakthroughs, for example, just on chronic neurological disorders could be life changing for the more than 10 million people worldwide—including 2 million Americans—who have been diagnosed with multiple sclerosis or Parkinson's plus the 40 million Alzheimer's patients (with 7 million Americans). Each year 10 million people will be told that they have a chronic neurological disorder.

**AI Milestone: AI Scientific Breakthroughs for the UN Sustainable Development Goals (SDGs).**[34] To encourage the greater use of AI systems in science and to help acknowledge the potential of AI systems to benefit humanity, this milestone is for an important scientific breakthrough using AI that helps with one or more of the UN SDGs.

**AI Milestone: Scientist's AI Aide/Collaborator.** John Jumper pointed out that one way to accelerate the pace of science is to improve the productivity of scientists. It's easy to imagine a GenAI aide that helps with grant writing and progress reports, which can be tedious. Another task would be to identify important new publications of interest to the scientist, ideally customized to the individual to summarize what is new compared to what the scientist already knew.[35] A more powerful AI system like the one Amodei envisions would go beyond what a Scientist's Aide could do. To keep the human involved to ensure safety and affordability, we pose this advance as a "scientist's AI collaborator."

[33] More powerful AI does not require achieving AGI, since it only needs to be an expert in one domain.

[34] This milestone is similar to No. four of the 10 Hard Problems in AI for 2050 [Manyika 2020].

# III. Harnessing AI for the Public Good

## Milestones, Prizes, and Research Centers

We proposed 18 AI milestones above where AI systems could benefit society. We next address how best to inspire and fund such AI research that enhances the public good. We recommend public and private interests pursue two paths to help researchers and practitioners reach these milestones.

The first is to create prizes each worth $1M+ to incentivize reaching milestones that significantly advance the positive impact of AI. Rather than recognize past achievements, *inducement prizes* try to stimulate research on focused targets. They can work well in many fields [Eisenstadt *et al*.] We plan to have inducement prizes for every milestone in this paper.

Table 2 lists 18 announced inducement prizes. Three of these prizes are ongoing, two were not awarded, and one was a partial award. The median time of the 12 awarded prizes is three years. The XPRIZE might be the best-known—a recent study of the XPRIZE touts its historical impact—but AI-related inducement prizes include the DARPA Self-Driving Grand Challenge in 2004, the Netflix Prize for user-rating prediction in 2009, and the 300+ ongoing Kaggle competitions. The US government itself supported more than 1,300 inducement prizes.

[35] The AI publication rate is so furious that some Berkeley grad students read 20 new papers per week just to keep up!

| Org | Competition Name | Given? | Start | Award |
|---|---|---|---|---|
| XPRIZE | Ansari: Launching a Space Industry | Yes | 1996 | 2005 |
| XPRIZE | Archon: Cheap, Accurate Genomics | No | 2006 to 2013 | — |
| XPRIZE | Progressive Insur-ance Automotive | Yes | 2007 | 2010 |
| XPRIZE | Google Lunar Landing | No | 2007 to 2018 | — |
| XPRIZE | Wendy Schmidt: Oil Cleanup | Yes | 2010 | 2011 |
| XPRIZE | Qualcomm: Tricorder | Partial Award | 2012 | 2017 |
| XPRIZE | Global Learning | Yes | 2014 | 2019 |
| XPRIZE | Shell: Ocean Discovery | Yes | 2015 | 2019 |
| XPRIZE | NRG COSIA: Carbon Reduction | Yes | 2015 | 2021 |
| XPRIZE | Anu & Naveen Jain: Women's Safety | Yes | 2016 | 2018 |
| XPRIZE | IBM Watson: AI & Global Challenges | Yes | 2018 | 2021 |
| XPRIZE | ANA: Robot Avatar | Yes | 2018 | 2022 |
| XPRIZE | Rainforest Understanding | Ongoing | 2019 | — |
| XPRIZE | Musk Foundation: Carbon Removal | Ongoing | 2021 | — |
| XPRIZE | Quantum Computing for Real-World Impact | Ongoing | 2024 | — |
| Netflix | Netflix Prize: Predict User Ratings | Yes | 2006 | 2009 |
| DARPA | DARPA Grand Challenge: Self Driving | Yes + Ongoing | 2004 | 2005+ |
| Kaggle | Kaggle: AI/ML Competitions | Yes + Ongoing | 2010 | 2011+ |

**Table 2. Outcomes of 18 example inducement prizes.**

For topics where there is not ample research, we recommend also funding ad hoc three- to five-year, high-impact, multidisciplinary research centers [Patterson].[36] Past examples were the UNIX project in the 1970s and the RADLab and Par Lab in the 2000s.[37] In addition to pursuing research that enables the milestones, centers might define success for the related prize, evolving the metrics and benchmarks used to show whether the milestone has been achieved so as to award the prize.

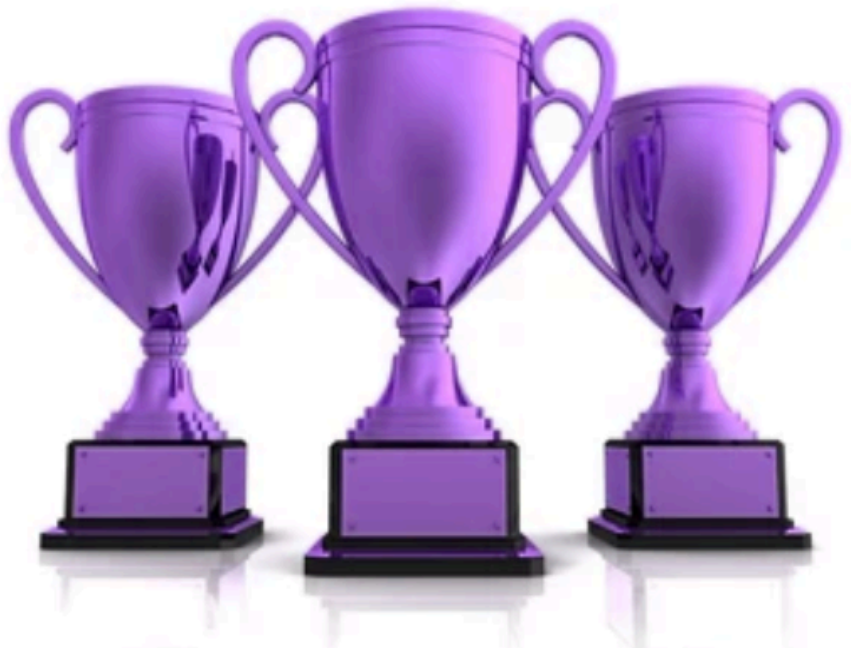

**Prizes can induce progress on AI milestones.**

We see the two paths as synergistic. Prize seekers can give feedback on the critical challenges in reaching a milestone for centers to pursue and give feedback on their progress; they can also offer technology transfer targets that research centers need to succeed.[38]

Research targets are not product announcements; we are not predicting that all 18 milestones will be realized in three to five years, but we expect many will. Table 2 demonstrates the high success rates of inducement prizes. Our aim is to give researchers targets that, if hit, would benefit society, with inducement prizes to inspire many to try.

The one-paragraph sketches of the proposed 18 milestones above are without much detail on what it would take to win the corresponding inducement prize. To flesh out what those details might look like, Appendix II gives an example of the details for the Rapid Upskilling prize from the Employment section.

# Conclusion

Several reports surveyed the state of the art of AI and considered the potential rewards and risks of AI [Bommasani *et al*.] [Brynjolfsson *et al*.] [Horvitz *et al.*] [Littman *et al*.] [Manyika 2022] [Reuel *et al*.]. Like some of these papers, rather than predict what the societal impact of AI systems *will be* assuming a laissez-faire approach, we envision what the impact *could be* given directed efforts in research and policy on using AI systems for good. We also propose 18 targets to show how to deliver on those efforts while thinking carefully about dissemination strategy and, as we shall see, funding.

While there are risks, there are also many known and unknown opportunities. It can be as big a mistake to ignore potential gains as it is to ignore risks. For example, some estimate that AI could plausibly raise the rate of growth in the U.S. gross domestic product from 1.4% currently to 3% [National Academies]. Doubling the growth rate could lead to "poverty reduction, better health care, improved environment, stronger national defense, and reduced budget deficit."

*It can be as big a mistake to ignore potential gains as it is to ignore risks.*

AI moves quickly, and governments must keep pace—or even better, ahead of developments. Decisions made today will shape the AI landscape of tomorrow, influencing everything from economic competitiveness to social stability. In the dawn of practical AI, thoughtful governance is not just desirable—it is essential.

Similar to how the government collaborated with industry in the successful development and deployment of cars and chips (see History), we encourage establishing a coordinated, public-private partnership for AI. It's important for industry players to participate to help build trust in AI systems with this wider audience. The goals of this partnership would be to remove bureaucratic roadblocks where possible, to ensure safety of the technology, and to educate and provide a transparent view of the development of this technology to both policymakers and the broader public.

[36] Interdisciplinary AI centers are also essential at the interface of AI and other domains. For example, the Bakar Institute for Digital Materials for the Planet's goal is to create a new integrated computational/experimental field of AI for materials to address climate change.

[37] The RADLab (Spark) and Par Lab (RISC-V) may not have as large an impact as UNIX, but they succeeded as multi-disciplinary research centers wth five-year sunset clauses.

[38] To ensure openness and transparency of research, we believe an organization that wins a research center should be ineligible for winning the associated research prize. A center's success should be measured on how well it helps others compete for the prize.

At this point, readers might expect that we scientists are about to ask for more government funding. We need government collaboration on the blueprint, but we believe that money for these efforts should come from the philanthropy of the technologists who have prospered in the computer industry. Several have already pledged support, and we expect more to join.

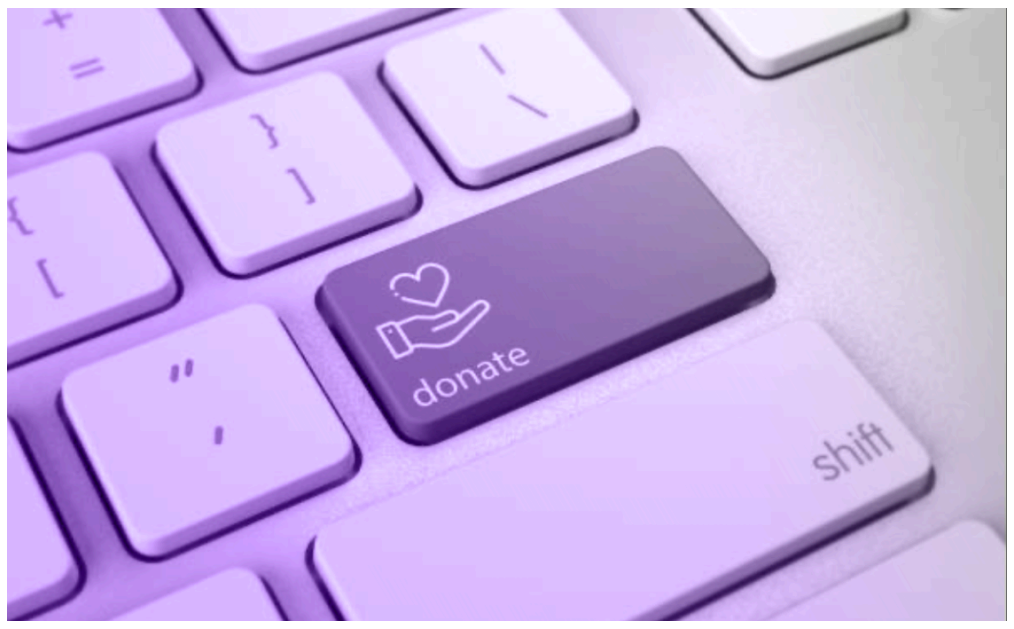

**Philanthropy should fund AI prizes and centers.**

One example is the *Laude Institute*, a soon-to-be-launched nonprofit organization committed to enabling computer science research aimed at benefiting society. All of its donations will come from people who have benefited financially from computer science research. Its focus is on shaping the global computing research agenda, catalyzing open-source communities, researching technology paradigm breakthroughs, and translating research into real-world solutions. One of its first tasks will be to use the funds raised to support many of the inducement prizes and research centers featured in this report.

As we wrapped up this effort, your authors spent more than a little time attempting to imagine the AI equivalent of President Kennedy's call in 1961 for going to the moon. It was a frustrating exercise. The fact is we can use AI to launch a thousand moonshots. We could aim to create an AI mediator that orchestrates conversations across political chasms to pull us out of polarization and back into pluralism. We can leverage the growing prevalence of smartphones by aiming to create a tutor app for every child in the world in their language, for their culture, and in their best learning style. We might enable biologists and neuroscientists to make a century of progress in a single decade. If we create the right innovation blueprint, we don't have to pick one moon.

---

*The fact is we can use AI to launch a thousand moonshots. ... If we create the right blueprint for innovation, we don't have to pick one moon.*

---

Our work is not done; we hope to recruit others to join the conversation. Achieving AI's potential in different domains will require a combination of domain experts, AI experts, policymakers, users of the technology in the domain, and other stakeholders to all come together to work toward solutions that empower people, bring new capabilities, and harness the power of AI for the public good.

# *Acknowledgements*

As scientists we collectively felt a sense of civic responsibility to communicate effectively with the public about an issue as important as AI. This obligation motivated the authors to volunteer a great deal of time for the past nine months on this report.

The views expressed herein are the authors' and do not necessarily reflect those of their employers, companies, and institutions: Alphabet, Carnegie Endowment for International Peace, Cornell University, Google, Harvard University, Lelepa AI, Perplexity, Stanford University, or UC Berkeley.

We wish to thank the topic experts for their valuable insights. For Employment, David Autor, Erik Brynjolfsson, Raluca Ada Popa, Tom Mitchell (who suggested the Job Forecaster); for Education, Salman Khan, Tom Kalil, Tom Mitchell, Caitlin Tenison; for Health, Adam Yala, David Feinberg, Robert Wachter; for Information / News / Social Networking, Dario Amodei, David Eaves, Krzysztof Gajos, Seth Lazar, Martin Wattenberg; for Governance / National Security, Jason Goldman, Barack Obama, Susan Rice, Eric Schmidt. For Media / Entertainment, Jon Finger, Alexandra Garfinkle, Maddy Buck, Neal Stephenson; and for Science, John Jumper, John Platt, Eric Schmidt.

Many gave us feedback that improved this paper, including Susan Athey, Jennifer Chayes, Marian Croak, Geoffrey Hinton, Samira Khan, Tom Kalil, Kurt Keutzer, Ed Lazowska, Derek Lockhart, Xiaoyu Ma, Christin Mills, James Manyika, Deirdre Mulligan, Raj Reddy, Daniel Rothchild, Alex Shammas, K. Tighe, Laura Tyson, and Cliff Young. We also acknowledge the helpful comments on our companion essay in *The Economist* from Scott Grafton, Scott Hubbard, Grace Mahusay, James Manyika, David Patterson Jr., Jessica Patterson, Jasper Rine, Susan Stafford, and K. Tighe.

# *Appendix I: Energy Usage of AI*

Electricity use is a portion of the global carbon footprint, alongside carbon fuel use by automobiles and planes, agriculture emissions, deforestation, and manufacturing goods. There are carbon dioxide equivalent emissions (CO2e) from manufacturing AI hardware and operating AI hardware, respectively called *embodied* CO2e and *operational* CO2e. We focus on operational CO2e as it is by far the largest piece.[39]

According to the International Energy Agency (IEA), data centers today use 1% of global electricity consumption, half of household digital electric appliances such as TVs, laptops, and cell phones [IEA].

[39] AI hardware lasts six or more years before replacement and data center buildings last decades.

Moreover, AI systems use a small portion of all data center electricity.[40] AI's share at Google was <15% each year from 2019 to 2021 [Patterson *et al*. 2022], which would put AI's share of global electricity consumption at <0.15%. That would mean digital home appliances consume more than 10 times as much electricity as AI does in data centers.[41]

AI relies on custom hardware that is five or more times as energy efficient for AI than conventional hardware. Hence, AI can be ≥80% of the data center computation but use ≤15% of its energy [Patterson *et al.* 2022]. If AI instead used standard computers, 80% of the computation would likely need more than half of the data center energy.

Besides data centers, smartphones also run AI. Smartphone electricity use is dominated by the display, wireless radio, and chargers, so AI uses <3% of smartphone electricity [Patterson *et al*. 2024].

The excitement about the potential of LLMs has inspired hyperscalers to make plans to expand their data centers at a time when many competing demands strain the present capacity of electric utilities. Several hyperscalers have announced goals of net zero emissions by 2030 to 2035, so they are investigating generating their own carbon-free energy, including nuclear energy, to be able to expand their data centers.

Data center growth is hard to predict accurately, but even if data center growth is strong, the IEA observes:

> *"However, when considered in a broader context of total electricity consumption growth globally, the contribution of data centres is modest. … continued economic growth, electric vehicles, air conditioners and the rising importance of electricity-intensive manufacturing are all bigger drivers."* [42] [IEA]

[40] AI also has already led to innovations that can reduce carbon. Examples are contrail reduction to reduce airline emissions (see Science), Project Green Light to minimize stop-and-go traffic and related emissions, and smart thermostats to reduce household energy use.

[41] Another perspective is that cryptocurrency consumes three to six times as much electricity as AI [Just Energy].

[42] Electricity-intensive manufacturing involves ferrous and non-ferrous metal (like aluminum) production when

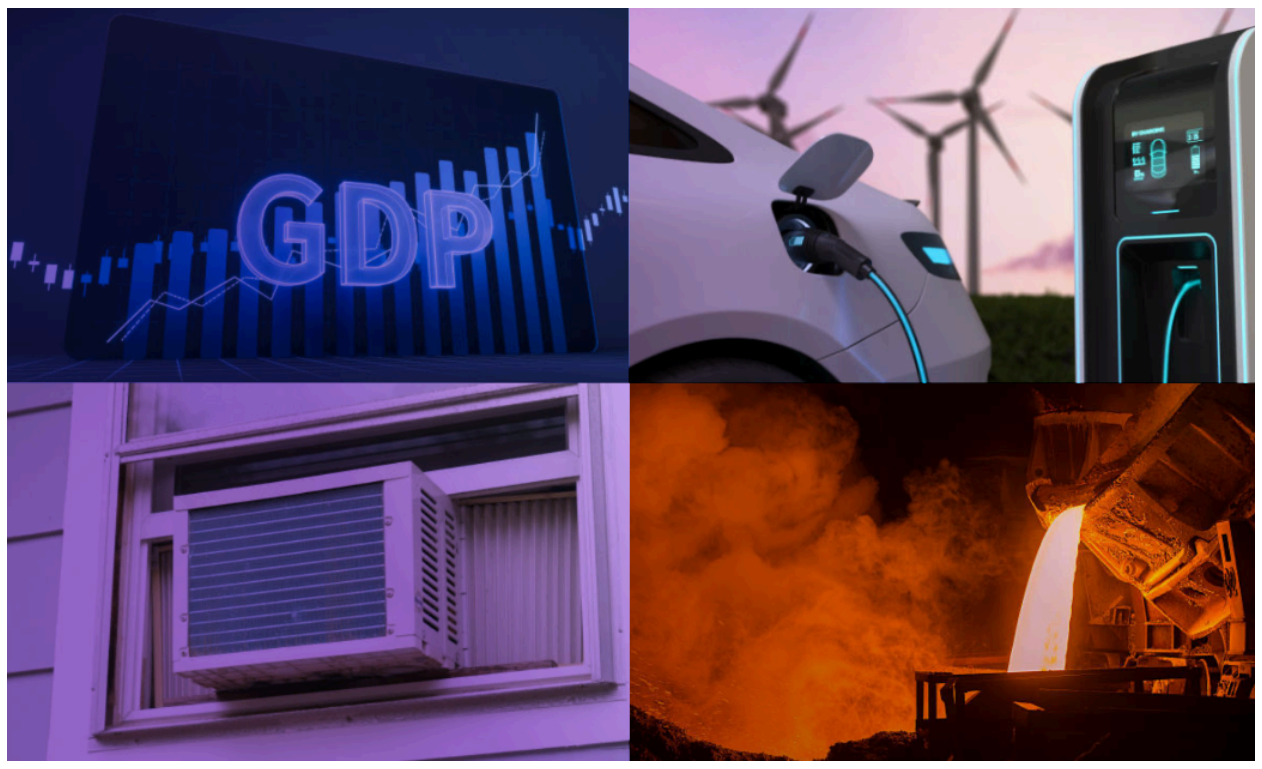

**AI is a small part of data center energy. Each use case will consume more electricity than data centers in 2030 [IEA] : continued economic growth, electric cars, air conditioners, and electricity-intensive manufacturing.**

While a modest global issue, adding or expanding many data centers in a single region can be problematic for a local utility.

# *Appendix II: The Rapid Upskilling Prize*

Above we give one-paragraph sketches of what the proposed 18 milestones are without much detail on what it would take to win the inducement prize. To flesh out what a prize might look like, this appendix gives an example of the details for one: the Rapid Upskilling milestone from the Employment Section.

The prize is $1M. Using the U.S. as a specific example, the goal is for full-time workers who earned ≤$30k per year to take a three- to six-month course that raises their income by ≥$15k per year after completing the course. (The U.S. federal poverty level for a family of four was $30,000 in 2023. The $30K and $15K targets might have to change elsewhere.)

Here are the requirements to win the prize:

- ≥50% of people who take the training complete the training (to be inclusive).
- ≥50% of people who complete the training increase their income for the subsequent year ≥$15k (to avoid cherry-picking).
- The program has at least 100 successful trainees (to demonstrate initial success).
- There must be vetted documentation of the trainees (e.g., pay stubs and W-2 forms in the U.S.) before and after the program (to ensure validity of claims).
- The program must document the potential scalability of the training system to hundreds of thousands or millions of people, and what the scaling would cost (to be able to address the size of the real problem).
- Prize submissions will be accepted on May 15 each year until the prize is awarded (to have sufficient time to collect prior evidence).

The judges may decide to offer an Honorable Mention Award of $100K to programs which, in their view, come close to the ultimate goal but do not yet fulfill all requirements of the Rapid Upscaling Prize.

---

done electrochemically, some clean energy technologies such as solar PV & batteries, and some electronic equipment.

# Authors

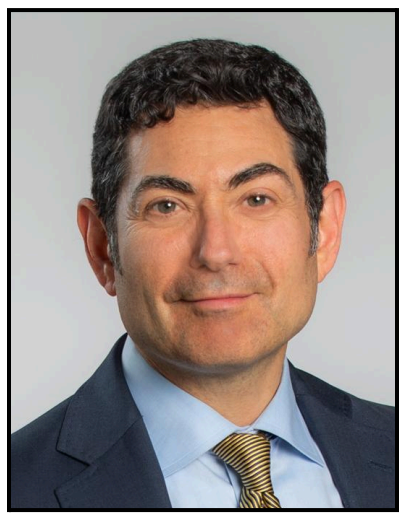

**Mariano-Florentino (Tino) Cuéllar**, former California Supreme Court justice and president of the Carnegie Endowment for International Peace, was the Stanley Morrison Professor at Stanford, chairs the Hewlett Foundation board, and is on the Inflection AI board.

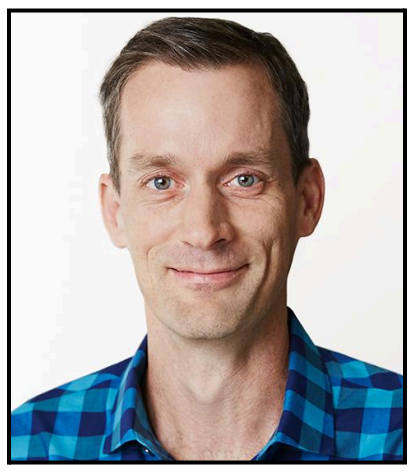

**Jeff Dean**, co-founder of Google Brain and Google Chief Scientist, shared the Association for Computing Machinery (ACM) Prize in Computing and received the IEEE John von Neumann medal.

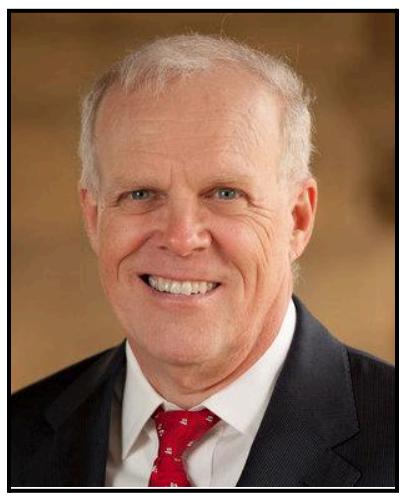

Professor **John Hennessy**, past Stanford University president, and Chairman of the Board of Alphabet, received the IEEE Medal of Honor and shared the ACM A.M. Turing Award and the National Academy of Engineering (NAE) Charles Draper Prize with David Patterson.

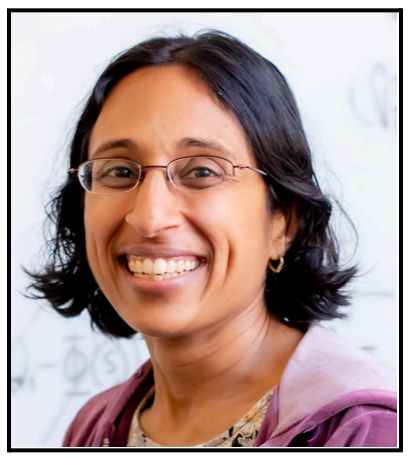

Harvard Professor **Finale Doshi-Velez** works on AI for healthcare. She was selected by the Institute of Electrical and Electronics Engineers (IEEE) as a "Top 10 in AI."

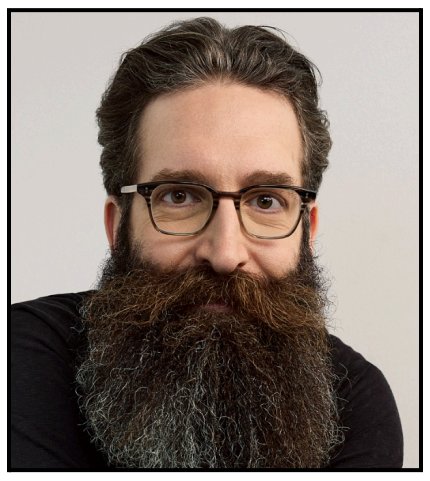

**Andy Konwinski** co-founded Databricks and Perplexity. During his Ph.D. studies at UC Berkeley he was a co-author with David Patterson on a vision paper that inspired this one.

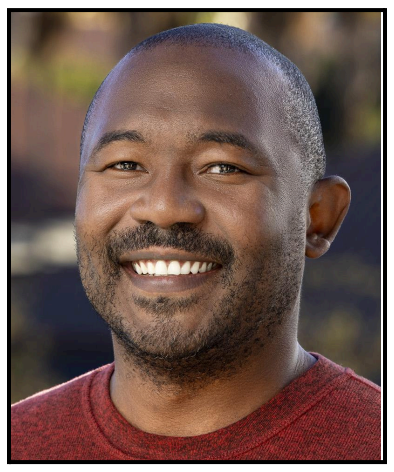

Professor **Sanmi Koyejo** works on trustworthy AI and is the current President of Black in AI. He won the 2021 Skip Ellis Early Career Award from the Computing Research Association (CRA).

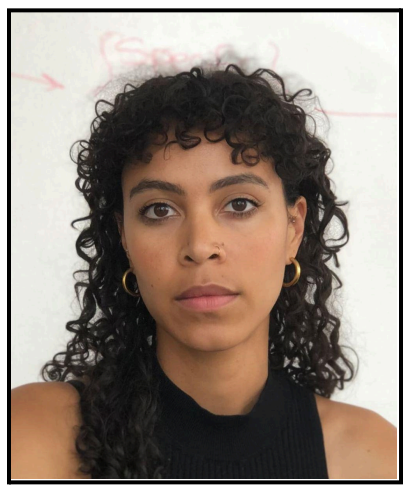

**Pelonomi Moiloa** is CEO of Lelepa AI (based in South Africa) and works on low-resource African languages. *Time* magazine named her one of the 100 most influential people in AI.

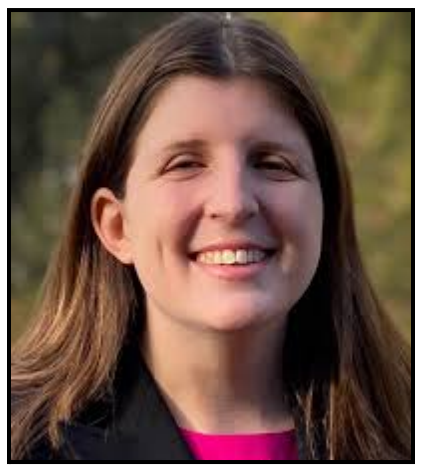

Cornell Professor **Emma Pierson** received a Rhodes Scholarship and was named to *Forbes'* "30 Under 30 in Science" and to *MIT Technology Review's* "35 Innovators Under 35."

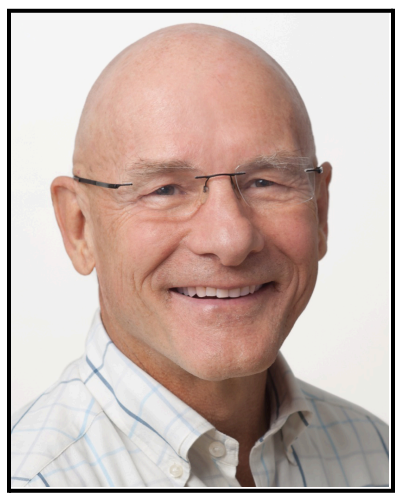

UC Berkeley Professor Emeritus **David Patterson**, past president of ACM, past chair of CRA, and Google Distinguished Engineer, shared the ACM A.M. Turing Award and the NAE Charles Draper Prize with John Hennessy.